\documentclass[10pt,twocolumn,letterpaper]{article}

\usepackage{cvpr}              % To produce the CAMERA-READY version
\usepackage[table]{xcolor}
\definecolor{cvprblue}{rgb}{0.21,0.49,0.74}
\usepackage[pagebackref,breaklinks,colorlinks,allcolors=cvprblue]{hyperref}
\usepackage{tikz}
\usetikzlibrary{positioning}

\def\paperID{*****} % *** Enter the Paper ID here
\def\confName{CVPR}
\def\confYear{2026}

\title{NTIRE 2026 Challenge on Efficient Low Light Image Enhancement: Methods and Results}

\author{
\scalebox{0.9}{ % 0.9 表示缩小到 90%，可以改为 0.8 等
\begin{tabular}{c} 
Jiebin Yan*\hspace{1.5em} Chenyu Tu*\hspace{1.5em} Qinghua Lin*\hspace{1.5em} Zongwei Wu*\hspace{1.5em} Weixia Zhang*\hspace{1.5em} Zhihua Wang*\hspace{1.5em} \\[2pt]
Peibei Cao*\hspace{1.5em} Yuming Fang*\hspace{1.5em} Xiaoning Liu*\hspace{1.5em} Zhuyun Zhou*\hspace{1.5em} Radu Timofte*\hspace{1.5em} Cheng Li\hspace{1.5em} \\[2pt]
Ziyi Wang\hspace{1.5em} Peishuai Zha\hspace{1.5em} Yuqiang Yang\hspace{1.5em} Jinao Song\hspace{1.5em} Guangsheng Tang\hspace{1.5em} Yan Chen\hspace{1.5em} \\[2pt]
Long Bao\hspace{1.5em} Heng Sun\hspace{1.5em} MD Raqib Khan\hspace{1.5em} Santosh Kumar Vipparthi\hspace{1.5em} Subrahmanyam Murala\hspace{1.5em} \\[2pt]
Shimon Murai\hspace{1.5em} Teppei Kurita\hspace{1.5em} Ryuta Satoh\hspace{1.5em} Yusuke Moriuchi\hspace{1.5em} Shang-Quan Sun\hspace{1.5em} \\[2pt]
Chenghao Qian\hspace{1.5em} Ruoyu Chen\hspace{1.5em} Zhirui Liu\hspace{1.5em} Wenqi Ren\hspace{1.5em} Haocheng Wu\hspace{1.5em} Changjian Wang\hspace{1.5em} \\[2pt]
Xuyao Deng\hspace{1.5em} Kele Xu\hspace{1.5em}  Hui Geng\hspace{1.5em} Qisheng Xu\hspace{1.5em} Zhidong Zhu\hspace{1.5em} Bangshu Xiong\hspace{1.5em}  \\[2pt]
Qiaofeng Ou\hspace{1.5em} Zhibo Rao\hspace{1.5em} Wei Li\hspace{1.5em} Xinbai Wang\hspace{1.5em} Duo Liu\hspace{1.5em} Chen Xie\hspace{1.5em} Jie Liu\hspace{1.5em} Jialu Xu\hspace{1.5em} \\[2pt]
Arya Shah\hspace{1.5em} Hriday Kamlesh Samdani\hspace{1.5em} Kishor Upla\hspace{1.5em} Kiran Raja\hspace{1.5em} Haodian Wang\hspace{1.5em} Zhi Jin\hspace{1.5em} \\[2pt]
Le Su\hspace{1.5em} Luwei Tu\hspace{1.5em} Fei Tan\hspace{1.5em} Jiawei Wu\hspace{1.5em} Hao Zhang\hspace{1.5em} Hongxi Huang\hspace{1.5em} Jinyi Xu\hspace{1.5em}  \\[2pt]
Pengfei Wei\hspace{1.5em} Yanhui Li\hspace{1.5em} Nikhil Akalwadi\hspace{1.5em} Harshitha M Banakar\hspace{1.5em} Sneha S Channappagoudar\hspace{1.5em} \\[2pt]
Spurti S M K\hspace{1.5em} Tejas Kumar M\hspace{1.5em} Ramesh Ashok Tabib\hspace{1.5em} Uma Mudenagudi\hspace{1.5em} Ju-Hyeon Nam\hspace{1.5em} \\[2pt]
Hyemin Shin\hspace{1.5em} Sang-Chul Lee\hspace{1.5em} Laura Álvarez-González\hspace{1.5em} Gibran Fuentes-Pineda\hspace{1.5em} \\[2pt]
Erik Molino-Minero-Re\hspace{1.5em} Juan C. Benito\hspace{1.5em} Marcos V. Conde\hspace{1.5em} Alvaro Garcia
\end{tabular}
}
}

\begin{document}
\maketitle

\begingroup
\renewcommand\thefootnote{\arabic{footnote}} 
\addtocounter{footnote}{-1}
\footnotetext{
    $^{*}$ J. Yan, C. Tu, Q. Lin, Z. Wu, W. Zhang, Z. Wang, P. Cao, Y. Fang, X. Liu, Z. Zhou and R. Timofte were the challenge organizers, while the other authors participated in the challenge. Each team described its own method in the report, shortened by the organizers to meet 8 page criteria. Appendix A contains the teams, affiliations and architectures if available. NTIRE 2026 webpage: \url{https://cvlai.net/ntire/2026/}.
}
\endgroup

\begin{abstract}
This paper presents a comprehensive review of the NITRE 2026 Efficient Low Light Image Enhancement (E-LLIE) Challenge, highlighting the proposed solutions and final outcomes.  This challenge focuses on mobile image enhancement under low-light conditions, aiming to design lightweight networks that improve enhancement quality while ensuring practical deployability under limited computational resources. A total of 207 participants registered, 27 teams submitted valid entries, and 17 teams ultimately provided valid factsheet. Based on these submissions, this paper provides a systematic evaluation of recent methods for E-LLIE, offering a comprehensive overview of state-of-the-art progress and demonstrating significant improvements in both performance and efficiency.  
\end{abstract}    
\section{Introduction}
\label{sec:intro}
Low-Light Image Enhancement (LLIE) aims to improve image visibility and contrast under various low-light conditions while addressing issues such as noise, artifacts, and color distortion commonly found in dark scenes. However, these degradation effects stem not only from the low-light environment itself but may also be further introduced during the brightness correction process.

In recent years, deep learning based methods for LLIE have achieved remarkable progress \cite{wang2025towards,wang2026robust,long2025enhancing}. However, these improvements are often accompanied by increased model complexity, leading to issues such as parameter redundancy, high computational cost, and long inference time, which hinder deployment on resource-constrained devices like mobile platforms. To address this challenge, Efficient Low-Light Image Enhancement (E-LLIE) has emerged, aiming to achieve high-quality enhancement with minimal computational overhead. In this context, model compression and acceleration techniques, including network pruning, quantization, neural architecture search, and knowledge distillation, have been widely explored to enable practical deployment.  

Building on the success of the NTIRE 2025 Low-Light Image Enhancement Challenge, we are launching the Efficient Low Light Image Enhancement Challenge at the NTIRE 2026 Workshop. This challenge continues to focus on efficient low-light enhancement and further encourages researchers to explore innovative solutions that balance enhancement quality and inference efficiency under strict computational resource constraints.

The goals of the challenge are threefold: (1) to advance research in low-light image enhancement under strict resource constraints,  (2) to facilitate systematic comparison of emerging methodologies; and (3) to provide a platform for academic and industrial participants to exchange ideas and explore potential collaborations. 

Unlike NTIRE 2025, the dataset used in this challenge was collected using smartphones from Huawei and Apple, and covers a variety of scenarios, including dim lighting, extremely low light, backlighting, uneven lighting, and indoor and outdoor night scenes. All images have a uniform resolution of 3024×4032. The dataset comprises 349 training scenes, 49 validation scenes, and 102 test scenes, with the ground truth (GT) images for the test set kept confidential from participants. Detailed specifications will be released at a later date. 

% This challenge is one of the NTIRE 2026 Workshop associated challenges on: **
This challenge is one of the challenges associated with the NTIRE 2026 Workshop
% ~\footnote{\url{https://www.cvlai.net/ntire/2026/}} 
on:
Deepfake detection~\cite{ntire26deepfake}, 
high-resolution depth~\cite{ntire26hrdepth},
multi-exposure image fusion~\cite{ntire26raim_fusion}, 
AI flash portrait~\cite{ntire26raim_portrait}, 
professional image quality assessment~\cite{ntire26raim_piqa},
light field super-resolution~\cite{ntire26lightsr},
3D content super-resolution~\cite{ntire263dsr},
bitstream-corrupted video restoration~\cite{ntire26videores},
X-AIGC quality assessment~\cite{ntire26XAIGCqa},
shadow removal~\cite{ntire26shadow},
ambient lighting normalization~\cite{ntire26lightnorm},
controllable Bokeh rendering~\cite{ntire26bokeh},
rip current detection and segmentation~\cite{ntire26ripdetseg},
low light image enhancement~\cite{ntire26llie},
high FPS video frame interpolation~\cite{ntire26highfps},
Night-time dehazing~\cite{ntire26nthaze,ntire26nthaze_rep},
learned ISP with unpaired data~\cite{ntire26isp},
short-form UGC video restoration~\cite{ntire26ugcvideo},
raindrop removal for dual-focused images~\cite{ntire26dual_focus},
image super-resolution (x4)~\cite{ntire26srx4},
photography retouching transfer~\cite{ntire26retouching},
mobile real-word super-resolution~\cite{ntire26rwsr},
remote sensing infrared super-resolution~\cite{ntire26rsirsr},
AI-Generated image detection~\cite{ntire26aigendet},
cross-domain few-shot object detection~\cite{ntire26cdfsod},
financial receipt restoration and reasoning~\cite{ntire26finrec},
real-world face restoration~\cite{ntire26faceres},
reflection removal~\cite{ntire26reflection},
anomaly detection of face enhancement~\cite{ntire26anomalydet},
video saliency prediction~\cite{ntire26videosal},
efficient super-resolution~\cite{ntire26effsr},
3d restoration and reconstruction in adverse conditions~\cite{ntire26realx3d},
image denoising~\cite{ntire26denoising},
blind computational aberration correction~\cite{ntire26aberration},
event-based image deblurring~\cite{ntire26eventblurr},
efficient burst HDR and restoration~\cite{ntire26bursthdr},
low-light enhancement: `twilight Cowboy'~\cite{ntire26twilight},
and efficient low light image enhancement.

\section{Tracks and Competition}
\label{sec:tracks}
\noindent\textbf{Ranking criteria.} 
To evaluate the submissions, we conducted a comprehensive evaluation of model performance using multiple metrics, including SSIM, LPIPS, DISTS, LIQE, and MUSIQ, with the additional constraint that the submitted model size is less than 1 MB. As show in Tab.~\ref{tab:submission_tab_final}, the "final ranking" was determined by aggregating the rankings of each team across all metrics to obtain an overall ranking. When methods had the same overall ranking, we further compared the number of model parameters; the method with fewer parameters received a higher final ranking. 

\noindent\textbf{Challenge phases.} 
(1) {\textit{Development and validation phase}}: Participants were provided with 349 training image pairs and 49 validation inputs from our custom dataset. The ground-truth images for the validation set were shared. Participants could submit their enhanced results to an evaluation server, which computed SSIM and LPIPS scores and provided real-time feedback. (2) {\textit{Testing phase}}: Participants received 102 low-light test images, without access to the corresponding ground-truth images. Submissions, including enhanced results, accompanying code, and a factsheet, were uploaded to the Codalab evaluation server and shared with the organizers. The organizers verified and the final results. Top-performing teams were required to submit training scripts to ensure reproducibility.

\begin{table*}[t]
\centering
\footnotesize
\setlength{\tabcolsep}{3pt}   % 减小列间距（关键）
\renewcommand{\arraystretch}{0.95} % 压缩行高（关键）

\caption{Evaluation and Rankings in the NTIRE 2026 Efficient Low Light Image Enhancement Challenge.}
\label{tab:submission_tab_final}

\resizebox{\textwidth}{!}{%
\begin{tabular}{ccccccccc}
\toprule
\textbf{Team} & \textbf{SSIM (Rank)} & \textbf{LPIPS (Rank)} & \textbf{DISTS (Rank)} & \textbf{LIQE (Rank)} & \textbf{MUSIQ (Rank)} & \textbf{Q-Align (Rank)} & \textbf{Params} & \textbf{Final Rank} \\
\midrule
MiVideo &0.5654 (8)&0.3632 (1)&0.1376 (1)&3.2561 (1)&68.8805 (1)&3.7699 (1)&927,049 &1\\[2pt]
CVPR TCD &0.5500 (14)&0.4801 (2)&0.2005 (3)&2.8661 (2)&63.9676 (5)&3.3778 (2)&557,618 &2\\[2pt]
S3 &0.5183 (16)&0.4157 (6)&0.2222 (2)&2.3865 (3)&64.9443 (2)&3.3718 (3)&741,600 &3\\[2pt]
sun &0.5688 (7)&0.4446 (9)&0.2231 (4)&2.2382 (7)&62.4497 (4)&3.2333 (4)&907,414 &4\\[2pt]
NCHU-CVLab &0.5557 (10)&0.5045 (5)&0.2542 (7)&2.3933 (4)&62.0570 (8)&3.3015 (10)&957,239 &5\\[2pt]
NUDT\_DeepIter &0.5211 (15)&0.5031 (4)&0.2317 (8)&2.4856 (6)&61.7693 (7)&3.2415 (5)&897,160 &6\\[2pt]
HIT-LLIE-team &0.5766 (4)&0.5176 (8)&0.2319 (11)&2.2977 (9)&60.3387 (12)&3.2007 (6)&101,922 &7\\[2pt]
VARCHASVI\_SVNIT &0.5575 (9)&0.4408 (17)&0.2414 (5)&1.5017 (12)&62.2430 (3)&3.0040 (8)&890,915 &8\\[2pt]
Xie\_Liu &0.5551 (11)&0.5171 (7)&0.2791 (6)&2.3798 (8)&62.0840 (10)&3.2130 (12)&913,388 &9\\[2pt]
JialuXu(IVC) &0.5122 (17)&0.5171 (3)&0.2430 (10)&2.6049 (5)&60.6554 (11)&3.2501 (9)&919,594 &10\\[2pt]
Bustaaa &0.5509 (12)&0.5143 (10)&0.2397 (9)&2.1875 (16)&61.2102 (9)&2.9807 (7)&205,361 &11\\[2pt]
sysu\_701 &0.5748 (5)&0.4828 (12)&0.2924 (14)&2.1568 (15)&56.7397 (6)&2.9837 (16)&965,254 &12\\[2pt]
SYSU-FVL\_ELLIE &0.5819 (1)&0.5561 (13)&0.2804 (15)&2.1494 (11)&55.9899 (15)&3.0264 (13)&994,871 &13\\[2pt]
KLETech-CEVI &0.5791 (2)&0.5894 (11)&0.3019 (13)&2.1737 (14)&57.5888 (17)&2.9884 (17)&525,429 &14\\[2pt]
ShinNam! &0.5789 (3)&0.5711 (15)&0.2677 (17)&2.0345 (13)&53.9563 (16)&2.9956 (11)&726,498 &15\\[2pt]
IIMAS-UNAM &0.5504 (13)&0.5536 (14)&0.2805 (12)&2.0970 (10)&57.9082 (14)&3.1247 (14)&890,915 &16\\[2pt]
Cidaut AI &0.5731 (6)&0.5282 (16)&0.2882 (16)&1.9627 (17)&54.5327 (13)&2.7156 (15)&797,222 &17\\
\bottomrule
\end{tabular}%
}
\end{table*}

\section{Challenge Methods and Teams}
\label{sec:method}
The results of the efficient low light enhancement challenge are detailed in Tab.~\ref{tab:submission_tab_final}, which evaluates and ranks the performances of 17 teams. 

The participating teams systematically explored a range of representative technical paradigms centered on the core objective of efficient low-light image enhancement. In general, these methods can be categorized into several main directions. First, lightweight network architectures based on U-Net or encoder–decoder frameworks are widely adopted, where techniques such as depthwise separable convolutions and residual connections are employed to reduce computational complexity while preserving representation capacity. Second, Retinex-based approaches or illumination–reflectance decomposition strategies are utilized to explicitly or implicitly model brightness and reflectance components, enabling targeted restoration of low-light degradation. Third, some methods incorporate lightweight attention mechanisms or Transformer modules to enhance global contextual modeling. In addition, several approaches leverage color space decoupling (e.g., HVI, YUV, LAB) to separately process luminance and chrominance information, thereby improving stability and color consistency. Meanwhile, certain methods integrate traditional image processing priors, such as histogram equalization and multi-exposure fusion, with deep learning frameworks to address limitations in extreme scenarios. Finally, to meet strict resource constraints, various efficiency-oriented techniques, including structural reparameterization and partial convolution, are extensively adopted to further improve inference efficiency. Overall, these approaches reflect a clear trend toward jointly optimizing enhancement performance and computational efficiency.

\subsection{MiVideo}
\begin{figure}[!htbp]
    \centering
    \includegraphics[width=8cm]{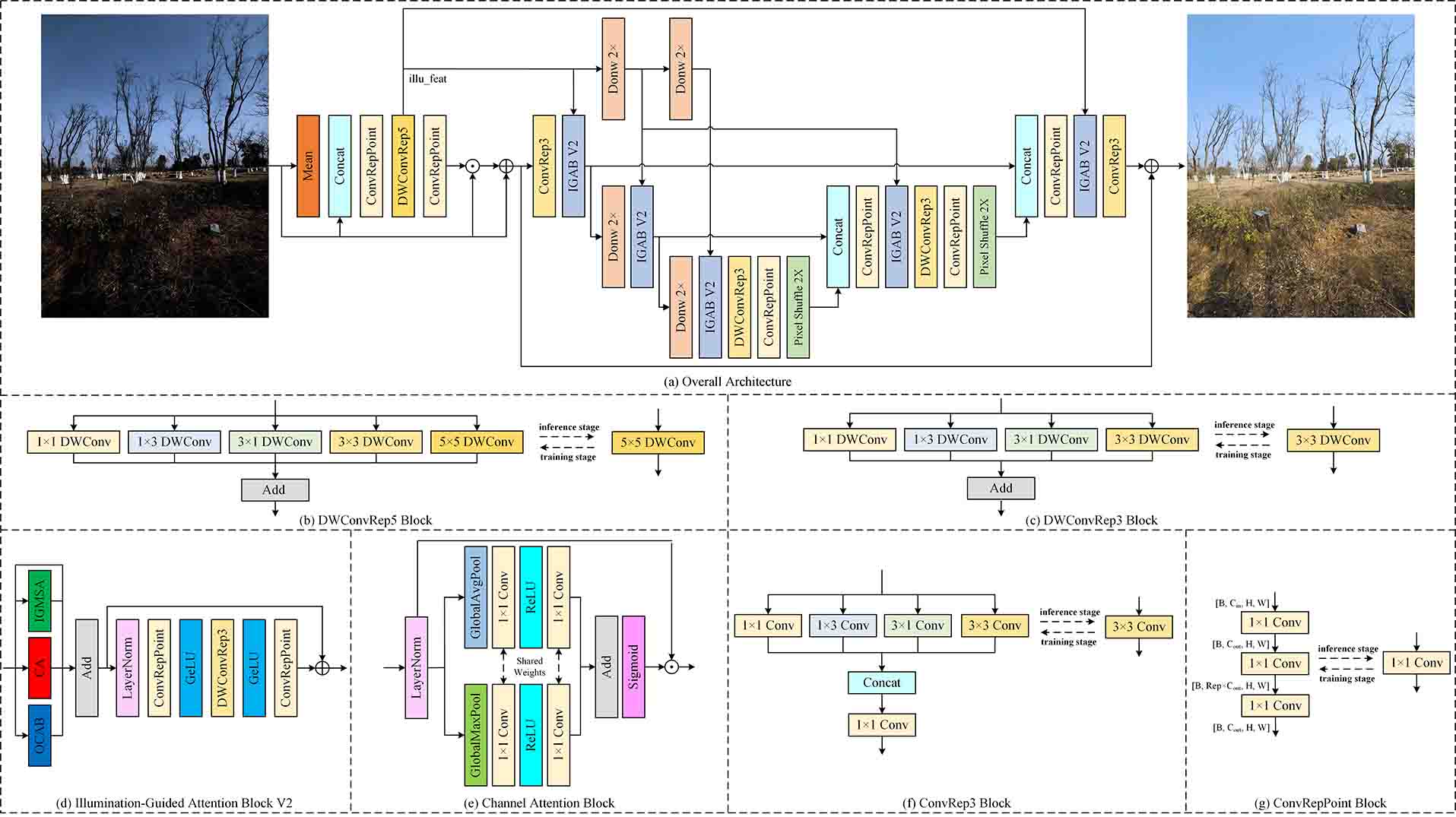}
    % \framebox(230,150){}
    \caption{The architecture of RetinexFormerRefine proposed by MiVideo.}
    \label{fig:RetinexFormerRefine}
\end{figure}
\noindent \textbf{Description:}
We adopt RetinexFormerRefine, a lightweight variant of RetinexFormer~\cite{cai2023retinexformer}, as our base model, as shown in Fig.~\ref{fig:RetinexFormerRefine}. The architecture follows the RetinexFormer design with a reduced channel size and a convolutional downsampling strategy. To enhance representation capability, we incorporate Channel Attention (CA) and OCAB modules into the original IGAB blocks, forming an improved IGAB V2. In addition, we employ structural reparameterization, which utilizes multi-branch structures during training and converts them into a single-branch structure during inference for higher efficiency.

\noindent \textbf{Implementation:} 
The model is implemented based on the BasicSR framework~\cite{basicsr} and trained on 8 NVIDIA A100 GPUs. We use the AdamW optimizer, cosine annealing learning rate scheduler, EMA, and learning rate warm-up. Training is conducted in three progressive stages with increasing input resolutions, and structural reparameterization is introduced in the final stage. The loss functions include L1, LPIPS, DISTS, SSIM, and Weighted TV. The training data is aligned using SuperPoint~\cite{detone2018superpoint} and LightGlue~\cite{lindenberger2023lightglue}, and a subset is reserved for validation. Training is iteration-based with early stopping, and the best-performing model on validation/test sets is selected.

% \subsection{XJRes (Excluded from Report)}

\subsection{CVPR TCD}
\begin{figure}[!htbp]
    \centering
    \includegraphics[width=8cm]{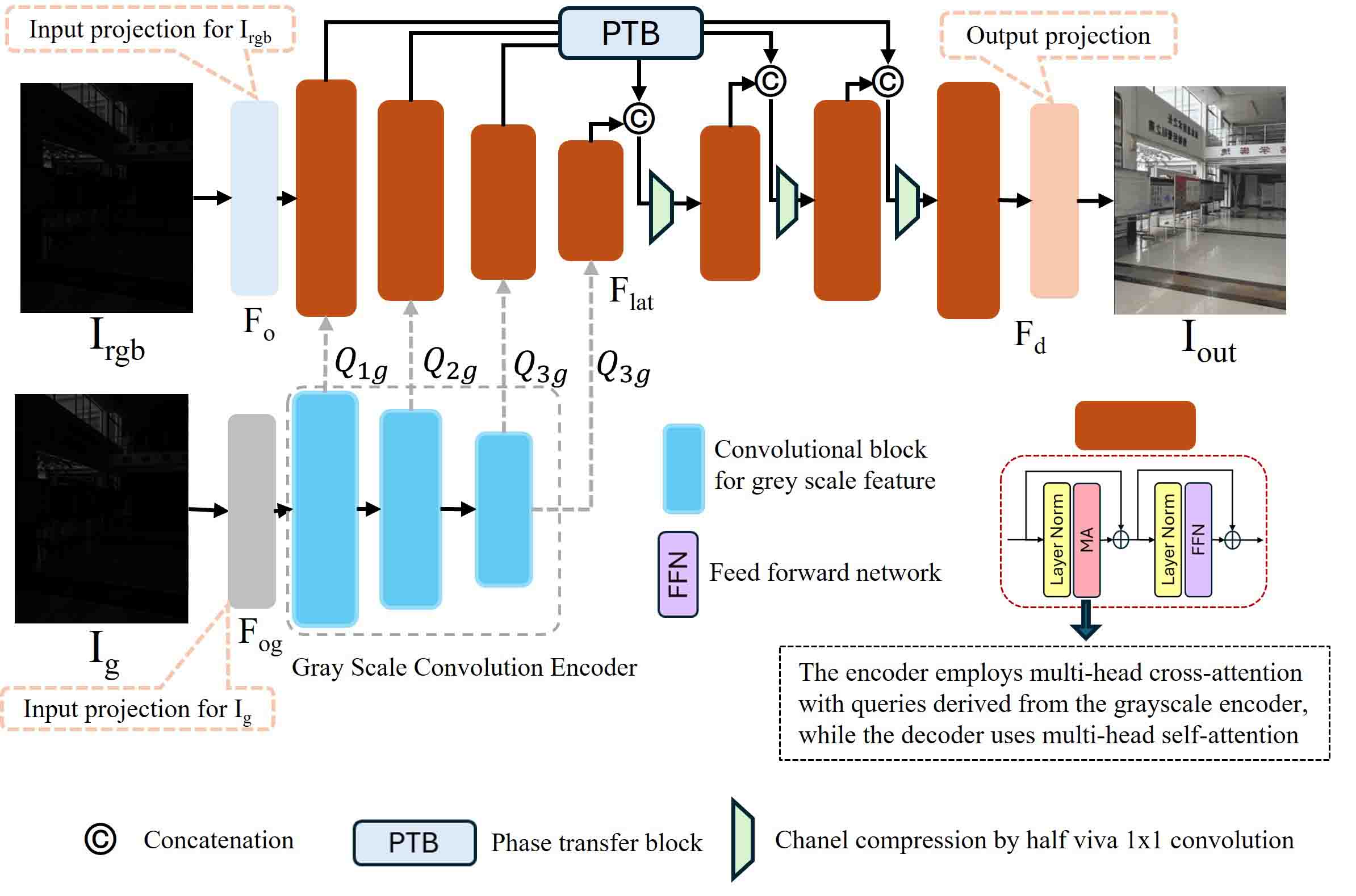}
    % \framebox(230,150){}
    \caption{Architecture of Team CVPR TCD}
    \label{fig:ntire2026LL}
\end{figure}

\noindent \textbf{Description:}
We propose a lightweight grayscale-guided cross-attention transformer network for efficient low-light image enhancement, as illustrated in Fig.~\ref{fig:ntire2026LL}. The model follows a dual-branch encoder--decoder architecture, where the RGB branch employs transformer blocks to capture global contextual information, while a parallel grayscale branch extracts structurally robust and illumination-invariant features using lightweight convolutional layers. A grayscale-guided cross-attention mechanism is introduced to fuse the two modalities, where grayscale features act as queries and RGB features serve as keys and values, enabling effective structure-aware enhancement. Furthermore, a Phase Transfer Block (PTB) is incorporated in the frequency domain to transfer phase information between encoder and decoder features, improving structural consistency and reconstruction quality. The overall design achieves a favorable balance between performance and efficiency, inspired by recent advances in transformer-based image restoration methods.

\noindent \textbf{Implementation:} 
The model is trained on a single NVIDIA GPU using the PyTorch framework. Training is conducted for 300 epochs with a batch size of 16. The optimization is performed using the Adam optimizer with an initial learning rate of $1 \times 10^{-4}$, which is gradually decayed during training. The loss function is defined as a combination of pixel-wise $\ell_1$ loss and perceptual loss to ensure both reconstruction accuracy and visual quality.

\subsection{S3}
\begin{figure}[!htbp]
    \centering
    \includegraphics[width=8cm]{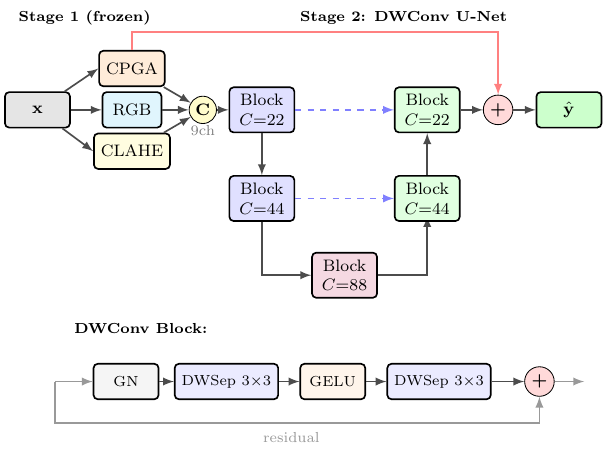}
    % \framebox(230,150){}
    \caption{Architecture of Team S3.}
    \label{fig:architecture_standalone}
\end{figure}
\noindent \textbf{Description:}
We follow a two-stage paradigm widely adopted in low-light image enhancement, similar to RetinexFormer~\cite{retinexformer} and HVI-CIDNet~\cite{hvicidnet}, consisting of a preprocessing module and a lightweight restoration backbone, as illustrated in Fig.~\ref{fig:architecture_standalone}. In the first stage, we employ two frozen algorithm-based preprocessors, CPGA-Net~\cite{cpganet} and CLAHE~\cite{clahe}, together with the original input, to perform preliminary brightness and contrast normalization. The outputs are concatenated into a 9-channel tensor, effectively reducing the distribution gap across diverse low-light inputs. In the second stage, we design a compact 3-level U-Net with depthwise-separable convolutions to minimize parameters while maintaining representation capacity. The network adopts an encoder–decoder structure with skip connections, GroupNorm, GELU activations, and residual learning, and incorporates a global residual connection from the CPGA-Net output to facilitate stable reconstruction.

\noindent \textbf{Implementation:} 
The model is trained for 500 epochs using AdamW optimizer~\cite{adamw} with a learning rate of $2 \times 10^{-4}$ and weight decay $1 \times 10^{-4}$, and a batch size of 16. A cosine annealing schedule with 10-epoch warmup is applied. The loss function combines pixel-wise and perceptual objectives, defined as $\mathcal{L} = 0.1 \cdot \mathcal{L}_{\ell_1} + \mathcal{L}_{\text{LPIPS}}$, where LPIPS measures perceptual similarity~\cite{lpips}. Training is conducted in full precision (fp32) on a single NVIDIA RTX PRO 5000 Blackwell GPU, with data augmentation including random $384 \times 384$ cropping and random horizontal/vertical flipping.

\subsection{sun}
\begin{figure}[!htbp]
    \centering
    \includegraphics[width=8cm]{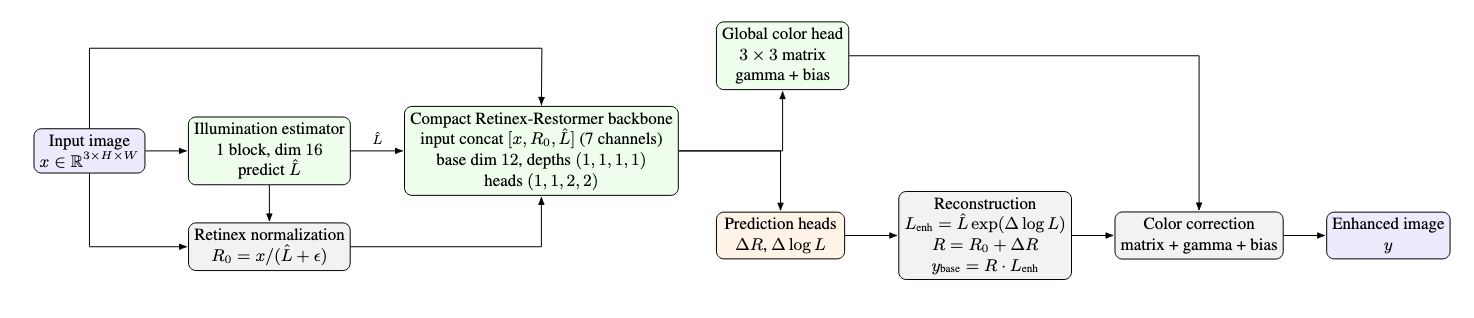}
    % \framebox(230,150){}
    \caption{Architecture of Team sun.}
    \label{fig:sun_arch}
\end{figure}
\noindent \textbf{Description:}
We adopt a compact Retinex-inspired encoder--decoder architecture augmented with Restormer-style local transformer blocks~\cite{sun2025di,sun2024restoring,zamir2022restormer,cai2023retinexformer}, as illustrated in Fig.~\ref{fig:sun_arch}. The model explicitly decomposes the input into illumination and reflectance components, where a lightweight illumination estimation branch guides the reflectance restoration process. To satisfy the strict 1\,MB model size constraint, both the channel width and network depth are carefully minimized, with shallow encoder/decoder stages and low-dimensional feature representations. The architecture maintains a balance between efficiency and performance by leveraging local attention mechanisms while avoiding heavy global modeling, and the final model is selected based on robustness across multiple evaluation metrics rather than a single objective.

\noindent \textbf{Implementation:} 
The model is trained for $40$ epochs using $768 \times 768$ random crops with a batch size of $1$ and gradient accumulation of $2$. The learning rate is set to $6 \times 10^{-5}$ with a $2$-epoch warm-up schedule. The training objective combines multiple loss components, including pixel-wise reconstruction (e.g., $L_1$), structural similarity, perceptual loss (LPIPS~\cite{zhang2018unreasonable} with AlexNet~\cite{krizhevsky2012imagenet}), and additional lightweight regularizations for color, luminance, and edge consistency, with weights such as $w_{\text{color}}=0.03$, $w_{\text{luma}}=0.15$, and $w_{\text{edge}}=0.05$. LPIPS is gradually introduced during the first $8$ epochs. Adversarial training is disabled in the final model due to instability under multi-metric optimization. Training is conducted on a single NVIDIA 3090 GPU, and the final checkpoint is selected based on a weighted combination of six evaluation metrics (SSIM, LPIPS, DISTS, LIQE, MUSIQ, and Q-Align).

% \subsection{NTR (Excluded from Report)}

\subsection{NCHU-CVLab}
\begin{figure}[!htbp]
    \centering
    \includegraphics[width=8cm]{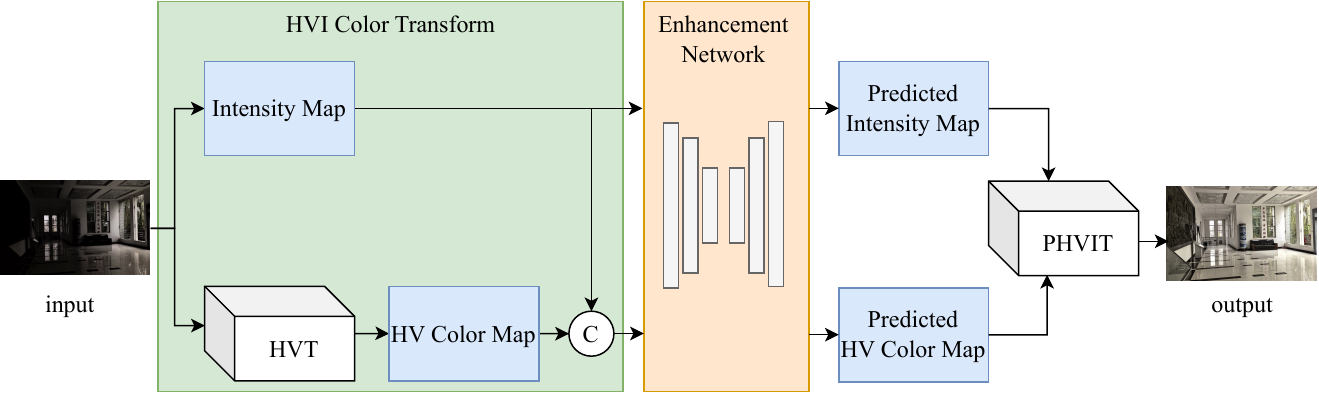}
    % \framebox(230,150){}
    \caption{Architecture of NCHU-CVLab.}
    \label{fig:NCHU-CVLab}
\end{figure}
\noindent \textbf{Description:}
We propose \textit{Efficient-HVI}, a lightweight low-light image enhancement network built upon the HVI color space~\cite{yan2025hvi,yan2024you}, as shown in Fig.~\ref{fig:NCHU-CVLab}. The method first transforms the input image into decoupled intensity and chromatic components, where the polarized hue-saturation representation alleviates red discontinuity while a learnable intensity collapse function suppresses noise in extremely dark regions. The network adopts a dual-branch architecture to separately model illumination adjustment and color restoration, and introduces cross-branch interaction to enhance feature fusion. This design enables efficient restoration with improved brightness consistency and color fidelity.

\noindent \textbf{Implementation:} 
The model is implemented in PyTorch and trained on four NVIDIA RTX 3090 GPUs. Training is conducted for 500 epochs using a batch size of 8 on $448 \times 448$ image patches. The initial learning rate is set to $1\times10^{-4}$ and scheduled with cosine restart, combined with a three-epoch warm-up strategy. The dataloader employs 16 worker threads with data shuffling, and gradient clipping is applied to stabilize training.

\subsection{NUDT\_DeepIter}
\begin{figure}[t]
\centering
\resizebox{\columnwidth}{!}{%
\begin{tikzpicture}[
    node distance=0.4cm,
    block/.style={
        rectangle, draw, fill=blue!20,
        text width=1.5cm,
        text centered,
        font=\scriptsize,
        minimum height=0.5cm
    },
    arrow/.style={->, >=stealth, thick},
    skip/.style={->, >=stealth, thick, dashed, red}
]
\node[block, fill=green!20] (in) {Input};
\node[block, right=of in] (e1) {Encoder1};
\node[block, below=of e1] (d1) {Down1};
\node[block, left=of d1] (e2) {Encoder2};
\node[block, below=of e2] (d2) {Down2};
\node[block, fill=orange!20, left=of d2] (bn) {Bottleneck};
\node[block, above=of bn] (u1) {Up1};
\node[block, right=of u1] (d3) {Decoder1};
\node[block, above=of d3] (u2) {Up2};
\node[block, right=of u2] (d4) {Decoder2};
\node[block, fill=green!20, right=of d4] (out) {Output};

\draw[arrow] (in) -- (e1);
\draw[arrow] (e1) -- (d1);
\draw[arrow] (d1) -- (e2);
\draw[arrow] (e2) -- (d2);
\draw[arrow] (d2) -- (bn);
\draw[arrow] (bn) -- (u1);
\draw[arrow] (u1) -- (d3);
\draw[arrow] (d3) -- (u2);
\draw[arrow] (u2) -- (d4);
\draw[arrow] (d4) -- (out);

\draw[skip] (e1) -- ++(0,0.6) -| (u1);
\draw[skip] (e2) -- ++(0,-0.6) -| (u2);
\end{tikzpicture}%
}
\caption{Architecture of Team NUDT\_DeepIter.}
\label{fig:architecture}
\end{figure}
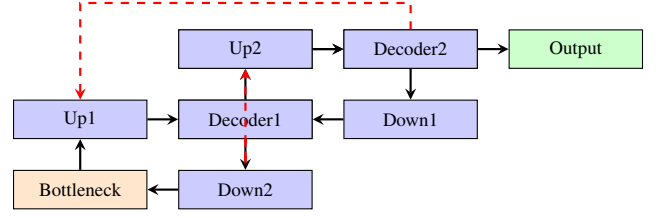

\noindent \textbf{Description:}
We propose \textbf{FinalEfficientEnhancerV2}, an efficient U-Net based architecture for low-light image enhancement following the encoder–decoder paradigm~\cite{retinexformer2023,uretinex2022}, as shown in Fig.~\ref{fig:architecture}. The network consists of three main components: an encoder with Partial Convolution (PConv) blocks for efficient feature extraction, a bottleneck with Dark Enhancement Blocks for adaptive low-light processing, and a decoder with skip connections to preserve high-frequency details. The PConv module, inspired by partial convolution~\cite{pconv2018}, processes only a subset of channels to significantly reduce computational cost. Additionally, Unified Inverted Bottleneck (UIB) blocks derived from MobileNetV3~\cite{mobilenetv32019} are used for lightweight design, while Dark Enhancement Blocks incorporate adaptive gating mechanisms similar to NAFNet~\cite{nafnet2022} to selectively enhance dark regions while maintaining well-lit areas.

\noindent \textbf{Implementation:} 
The model is implemented in PyTorch (Python 3.8) and trained for 600 epochs on the LOL dataset~\cite{lol2018}. We adopt the AdamW optimizer with an initial learning rate of $5\times10^{-5}$ and apply cosine annealing scheduling. The training objective is a weighted combination of multiple loss functions, including MS-SSIM, L1, perceptual, dark region, color, frequency, and SNR losses. The model is designed to be lightweight and efficient, containing approximately 450K parameters, enabling fast inference while maintaining high enhancement quality.

\subsection{HIT-LLIE-team}
\noindent\textbf{Description:}
We propose an ultra-lightweight fully convolutional network, termed MobileIE-6Ch, for efficient low-light image enhancement, built upon the MobileIE baseline~\cite{MobileIE}. The architecture introduces three key components: (1) a unified 6-channel output head that jointly models RGB illumination and noise residual, (2) a Multi-Branch Reparameterization (MBR) backbone for enhanced feature extraction, and (3) a dual-attention modulation module for adaptive feature refinement. During training, multi-branch convolutions (e.g., $5\times5$, $3\times3$, $1\times1$, and asymmetric kernels) are employed and later structurally reparameterized into single convolutions for inference, following~\cite{ding2021repvgg}, enabling a highly compact model with negligible inference overhead. The network follows a Retinex-based formulation~\cite{retinex_theory}, where the enhanced image is reconstructed via predicted illumination and residual components, allowing simultaneous brightness enhancement and noise suppression.

\noindent\textbf{Implementation:}
The model is trained using the Adam optimizer~\cite{kingma2014adam} with a batch size of 16 for 800 epochs under a Distributed Data Parallel (DDP) setting. Training is divided into a 20-epoch warm-up stage (learning rate $1\times10^{-6}$) and a main training stage (initial learning rate $1.5\times10^{-4}$) with cosine annealing scheduling~\cite{loshchilov2016sgdr}. A hybrid loss function is adopted, combining Charbonnier loss~\cite{charbonnier1994two}, SSIM loss~\cite{wang2004image}, color consistency, and frequency-domain constraints. High-resolution images are processed using random cropping ($768\times768$ patches), and a 5-fold cross-validation strategy is applied. Additional stabilization techniques include gradient clipping (threshold 0.5) and careful illumination regularization during warm-up.

\subsection{VARCHASVI\_SVNIT}

\noindent \textbf{Description:}
We propose \textbf{SCALENet}, an illumination-aware deep network for low-light image enhancement that dynamically adapts its behavior based on the input luminance, as shown in Fig.~\ref{fig:VARCHASVI_SVNIT}. The model is built around three key components. First, an \emph{Illumination-Conditioned Normalisation (ICN)} module replaces standard normalization with a FiLM-inspired conditioning mechanism, where a global luminance scalar modulates channel-wise affine parameters, enabling adaptive feature scaling across varying lighting conditions. Second, a \emph{Multi-Scale Exposure Fusion (MSEF)} module processes three virtual exposures (under-, normal-, and over-exposed) through parallel CNN branches and fuses them using a spatially-adaptive softmax weighting conditioned on the luminance map. Third, a \emph{Deep Refinement Head} with spatial and channel attention further enhances details, where spatial attention focuses on darker regions and channel attention corrects residual color biases. This design allows SCALENet to effectively restore perceptual quality while remaining lightweight and adaptive to scene illumination.

\noindent \textbf{Implementation:}
The model is trained using a composite loss function that jointly optimizes multiple perceptual and structural metrics, including MS-SSIM, VGG-based perceptual loss, LPIPS, FFT-based frequency loss, color constancy losses, and gradient loss, with carefully balanced weights. Training is performed using the AdamW optimizer with an initial learning rate of $2\times10^{-4}$, weight decay $10^{-4}$, and cosine annealing down to $10^{-6}$, along with gradient clipping at 1.0. An exponential moving average (EMA) with decay 0.999 is maintained for stable inference. The training follows a progressive resolution strategy: $0$--$39$ epochs with $256\times256$ crops and batch size $16$, $40$--$89$ epochs with $384\times384$ crops and batch size $8$, and $90+$ epochs with $512\times512$ crops and batch size $4$, with learning rate halved at each stage transition. Additional strategies include illumination-aware data sampling to rebalance dark images and standard augmentations such as flipping, rotation, and cropping.

\subsection{Xie\_Liu}
\noindent \textbf{Description:}
We propose HE-Net, an efficient framework for low-light image enhancement that combines histogram equalization (HE) with a progressive CNN-based refinement strategy, as shown in Fig.~\ref{fig:xieliu}. Specifically, HE is first applied to low-resolution inputs to improve contrast while mitigating noise amplification \cite{gonzalez2009digital}. The enhanced representations are then progressively reconstructed to higher resolutions through a multi-stage architecture, where each stage employs residual blocks to fuse upsampled images and learned feature maps. A key component is a differentiable gamma correction module that predicts pixel-wise gamma values for adaptive enhancement. Feature extraction is conducted via an Inception-like design that aggregates multi-scale depth-wise convolutions and global context \cite{szegedy2015going}. For color images, enhancement is performed in the LAB space by refining the luminance channel and reconstructing chrominance via lightweight residual learning, enabling effective and efficient restoration.

\noindent \textbf{Implementation:}
The model is implemented in PyTorch and trained end-to-end on multiple datasets, including LOLv2 \cite{yang2021sparse}, LSRW \cite{hai2023r2rnet}, and the NTIRE-2026 training set, with weighted sampling. Training is conducted on two NVIDIA TITAN RTX GPUs using FP16 precision. Images are resized to $480 \times 640$ for efficiency. Optimization is performed using AdamW with L2 regularization, an initial learning rate of $1\times10^{-4}$, and scheduled decay at the 10th and 15th epochs. The network is trained for 20 epochs with a batch size of 8 per GPU. The loss function is a weighted combination of L1 loss, SSIM loss, and LPIPS loss.

\subsection{JialuXu(IVC)}
\noindent \textbf{Description:}
We propose a lightweight low-light image enhancement (LLIE) framework that integrates both global priors and efficient high-resolution restoration, as shown in Fig.~\ref{fig:JialuXu}. Specifically, a channel-wise histogram prior learned from paired data provides stable global illumination initialization, while an image-adaptive dynamic prior branch predicts compact modulation parameters conditioned on the input. The main backbone operates in a space-to-depth domain via \texttt{PixelUnshuffle}, enabling a larger effective receptive field under strict efficiency constraints. To further enhance contextual modeling, multi-stage depthwise dilated convolutions are employed to capture long-range dependencies in challenging low-light regions. In addition, a local color correction matrix (CCM) head performs spatially varying cross-channel calibration, improving color fidelity and perceptual realism. This design effectively balances restoration quality and computational efficiency, achieving competitive perceptual performance as validated by no-reference metrics such as LIQE~\cite{Zhang_2023_CVPR} and Q-Align~\cite{wu2024qalign}.

\noindent \textbf{Implementation:}
The model is trained using only the official dataset, with a selected subset of 500 image pairs derived from limited augmentation. Training is conducted on an NVIDIA RTX 3060 GPU (12 GB memory) using mixed-precision. We adopt a batch size of 4 and train for 16 epochs. Optimization is performed with AdamW, using an initial learning rate of $8 \times 10^{-4}$, weight decay of $10^{-4}$, and cosine annealing scheduling. The training objective combines multiple loss functions, including Charbonnier loss, L1 loss, color-consistency loss, gradient loss, exposure loss, and saturation loss. The model contains 222{,}078 trainable parameters and maintains a sub-megabyte checkpoint size, ensuring efficiency for deployment under resource-constrained environments.

% \subsection{UPT-eLLIE (Excluded from Report)}

\subsection{Bustaaa}
\noindent \textbf{Description:}
We propose a lightweight dual-branch network for low-light image enhancement based on LYT-Net~\cite{lytNet}, as shown in Fig.~\ref{fig:Bustaaa}. The model operates in the YUV color space, where luminance and chrominance components are processed separately. The luminance branch enhances the $Y$ channel using convolution, pooling, and Multi-Head Self-Attention (MHSA) to capture global illumination-aware features. Meanwhile, the chrominance branch refines the $U$ and $V$ channels through a channel-wise refinement module to suppress chroma noise while preserving color fidelity. The two branches are fused via a lightweight module with depthwise convolutions and adaptive channel reweighting, followed by reconstruction to generate the final RGB image. The training objective adopts a hybrid loss combining Smooth L1 loss, perceptual loss~\cite{perceptual_loss}, histogram loss~\cite{HE_loss}, color-consistency loss~\cite{color_loss}, and MS-SSIM loss~\cite{ssim_loss}.

\noindent \textbf{Implementation:}
The model is implemented in PyTorch and trained end-to-end using the Adam optimizer. The initial learning rate is set to $1\times10^{-4}$ with a cosine annealing schedule. Training is conducted on randomly cropped aligned patches of size $512 \times 512$, with random flipping and rotation for data augmentation. During inference, high-resolution images are processed in a patch-wise manner using the same crop size, and the restored patches are stitched to obtain the final output. The model is trained only on the official paired dataset without external data or pretraining.

\subsection{sysu\_701}
\noindent \textbf{Description:}
We propose an improved and lightweight low-light image enhancement model based on the HVI-CIDNet architecture, as shown in Fig.~\ref{fig:sysu701}, which leverages the Horizontal/Vertical-Intensity (HVI) color space to decouple chromatic and intensity information and mitigate noise issues inherent in traditional color spaces~\cite{yan2025hvinewcolorspace}. The model adopts a dual-branch structure consisting of an HV branch for color representation and denoising, and an I branch for illumination enhancement, where cross-attention (LCA) facilitates information exchange between the two branches. To improve efficiency, we redesign the backbone by reducing the U-Net depth from five to four layers, compressing channel dimensions, and significantly decreasing the number of attention heads. In addition, we enhance the loss formulation by combining Charbonnier loss, multi-scale SSIM, color constancy loss, and frequency-domain loss to better preserve structural fidelity, color consistency, and fine texture details.

\noindent \textbf{Implementation:}
The model is implemented in PyTorch and trained on an NVIDIA A100 GPU. Training is conducted for 1000 epochs with a batch size of 8 using the Adam optimizer, with an initial learning rate of $1 \times 10^{-4}$. A Cosine Annealing Restart scheduler is applied, where the minimum learning rate is set to $1 \times 10^{-7}$ to ensure stable convergence. During preprocessing, input images are randomly cropped to $256 \times 256$ patches. Additional training strategies include dynamic HVI weighting, gamma enhancement, and gradient clipping to stabilize optimization and improve visual quality.

\subsection{SYSU-FVL\_ELLIE}
\noindent \textbf{Description:}
We adopt a lightweight encoder--decoder architecture with skip connections as the backbone, where the encoder is built upon depthwise separable convolutions (DWS)~\cite{Mobilenets} and employs a larger $5\times5$ kernel in the bottleneck to enlarge the receptive field. In the decoder, a HybridBlock is introduced at the bottleneck, which combines multi-scale DWS convolutions ($3\times3$ and $5\times5$) with lightweight channel attention, forming a CNN--attention hybrid design that captures both local and global contextual information efficiently. Additionally, a ColorCorrectionTail module is appended to mitigate color artifacts. The network produces multi-scale outputs at $1\times$, $1/2\times$, and $1/4\times$ resolutions to facilitate robust reconstruction.

\noindent \textbf{Implementation:}
The model is implemented in PyTorch and trained on an NVIDIA A6000 GPU. The total loss consists of an $L_1$ loss, SSIM loss, perceptual loss based on a pretrained VGG19 network~\cite{vgg19}, least-squares GAN loss~\cite{lsgan}, and feature matching loss, with respective weights $\lambda_s=0.2$, $\lambda_p=0.04$, $\lambda_g=0.05$, and $\lambda_{fm}=5.0$. Optimization is performed using Adam~\cite{adam} with an initial learning rate of $2\times10^{-4}$, scheduled by cyclic cosine annealing~\cite{cosine}. The model is trained from scratch for 150{,}000 iterations with a batch size of 4 and patch size of $2048\times2048$, while GAN training is activated after 10{,}000 iterations.

% \subsection{UNet\_YYDS (Excluded from Report)}

\subsection{KLETech-CEVI}
\noindent \textbf{Description:}
We propose \textbf{ResLNet}, a lightweight fully convolutional residual network designed for low-light image enhancement, as shown in Fig.~\ref{fig:KLETech-CEVI}. The model directly maps a low-light RGB image to an enhanced RGB image at the same spatial resolution, avoiding any downsampling or upsampling operations to preserve fine-grained details. The architecture consists of four main components: a convolutional head, a residual body with 12 residual blocks, a Global Illumination Module (GIM), and a reconstruction tail. The residual blocks employ depthwise separable and dilated convolutions together with channel attention mechanisms to efficiently capture contextual information while maintaining a compact parameter footprint. The network adopts residual image learning with a sigmoid activation at the output stage to constrain predictions within a valid intensity range. This design ensures both computational efficiency and high-quality enhancement, making it suitable for resource-constrained deployment.

\noindent \textbf{Implementation:}
The model is trained using paired low-light and normal-light images provided by the NTIRE 2026 ELLIE Challenge, with a 90\%/10\% train-validation split. Training is implemented in PyTorch with all images preloaded into memory to reduce I/O overhead, and standard data augmentations including random cropping and flipping are applied. The network is trained for 150 epochs with a batch size of 16. The optimization uses AdamW with a learning rate of $2\times10^{-3}$ and weight decay of $10^{-2}$, together with a cosine annealing scheduler ($T_{\max}=120$, $\eta_{\min}=10^{-6}$). The loss function is a weighted combination of SSIM, L1, and gradient losses, defined as $\mathcal{L} = 0.75\,\mathcal{L}_{\mathrm{SSIM}} + 0.20\,\mathcal{L}_{1} + 0.05\,\mathcal{L}_{\mathrm{grad}}$. Model selection is based on validation SSIM, and training utilizes the \texttt{piq} library for differentiable SSIM computation.

\subsection{ShinNam!}
\noindent \textbf{Description:}
Our solution adopts a lightweight blind low-light image enhancement and restoration network designed for real-world all-in-one image restoration, as shown in Fig.~\ref{fig:ShinNam!}. The architecture follows a compact U-Net~\cite{ronneberger2015u} style encoder--decoder framework with a context-adaptive bottleneck. Specifically, a global degradation context vector is first extracted from the input image via a dedicated context extraction branch. This context is then used to modulate bottleneck features through adaptive residual blocks, enabling dynamic feature adaptation according to degradation characteristics. The decoder reconstructs the restored image with skip connections, and the final output is generated in a residual learning manner by adding the predicted residual to the input followed by a sigmoid activation.

\noindent \textbf{Implementation:}
The model is trained using the Adam optimizer with a learning rate of $1\times10^{-4}$, a batch size of 8, and 100 training epochs. Input images are resized to $512\times512$. The training data consists of the official paired dataset along with additional patch data generated a four-way split strategy. The loss function is a combination of Charbonnier reconstruction loss and spatial consistency loss, where the former enforces pixel-wise fidelity and the latter preserves local structural relationships.

\subsection{IIMAS-UNAM}
\noindent \textbf{Description:}
We adopt a compact enhancement framework termed \textbf{Hierarchical Global Masking Network (HGM-Net)}, built upon a lightweight U-Net-like encoder--decoder architecture~\cite{ronneberger2015unet}, as shown in Fig.~\ref{fig:IIMAS-UNAM}. The model introduces a hierarchical routing strategy that decomposes enhancement into global scene-level decisions and local region-aware corrections. The backbone consists of a shallow encoder--decoder with progressively increasing channels, integrating depthwise residual bottleneck blocks, skip connections, squeeze-and-excitation modules~\cite{hu2018senet}, and a global-context block~\cite{cao2019gcnet} to capture both local and global dependencies. Additionally, the network employs multiple exposure-aware masks to adaptively modulate illumination and structural attributes across different regions. A dedicated detail branch and a Mamba-inspired color modeling branch~\cite{gu2024mamba} are fused to generate the final enhanced output, enabling efficient yet expressive low-light image enhancement.

\noindent \textbf{Implementation:}
The model is trained from scratch using only the official challenge dataset. Training is performed with random initialization using a batch size of 4, a crop size of $384\times384$, and a learning rate of $1\times10^{-4}$ under mixed-precision settings. The training process runs for 32 epochs. The optimization objective combines multiple loss terms, including pixel-wise reconstruction loss, SSIM and MS-SSIM, perceptual loss (LPIPS-style), gradient regularization, and auxiliary routing supervision. During inference, images are padded to match architectural constraints, processed in a single forward pass, and restored to their original resolution without any test-time augmentation or ensemble strategy.

% \subsection{CVIR\_Lab (Excluded from Report)}

% \subsection{SDUMVP-LLIE (Excluded from Report)}

% \subsection{PALLab (Excluded from Report)}

% \subsection{Conqueror (Excluded from Report)}

\subsection{Cidaut AI}
\noindent \textbf{Description:}
We adopt a modified CIDNet architecture~\cite{yan2024you} for efficient low-light image enhancement, as shown in Fig.~\ref{fig:Cidaut AI}. The original CIDNet consists of multiple Lighten Cross-Attention (LCA) blocks, where each block sequentially applies a Cross Attention Block (CAB) and an Image Enhancement Layer (IEL). The CAB facilitates information exchange between hue-value and intensity features in the HVI color space, while the IEL enhances illumination. In our approach, we replace the IEL with a novel NIEL block, which is built upon the NAFBlock design~\cite{chen2022simple} and further inspired by FIE-Block and the original IEL. The NIEL block incorporates multiple depth-wise convolutions for efficiency, along with Simple Gate (SG) and Simplified Channel Attention (SCA) modules to improve feature representation while maintaining low computational cost.

\noindent \textbf{Implementation:}
The model is implemented in PyTorch~\cite{paszke2019pytorch}. We adopt a lightweight channel configuration of [6,6,8,16] to reduce model complexity. The network is trained using the Adam optimizer~\cite{diederik2014adam} with a learning rate of $1\times10^{-4}$, weight decay of $1\times10^{-3}$, and a cosine annealing schedule down to $1\times10^{-7}$. Training is conducted for 800 epochs with a batch size of 8 on a single NVIDIA H100 GPU. The loss function follows CIDNet~\cite{yan2024you}, combining RGB and HVI losses, each composed of L1, SSIM, edge, and perceptual terms with predefined weights.

% \subsection{cvLab (Excluded from Report)}

% \subsection{Amrita LLIR (Excluded from Ranking)}

\section*{Acknowledgements}
This work was partially supported by the National Natural Science Foundation of China (Grants 62371283, 62461028, 62301323, and 62501293) and the Humboldt Foundation. We thank the NTIRE 2026 sponsors: OPPO, Kuaishou, and University of Wurzburg (Computer
Vision Lab).

{
    \small
    \bibliographystyle{ieeenat_fullname}
    \bibliography{main}
}


\clearpage 
\appendix  

\section{Teams and Affiliations}
\label{sec:appendix_teams}

% --- 团队 1：NTIRE 2026 team (组织者) ---
% --- 团队 1：NTIRE 2026 team (组织者) ---
\noindent \textbf{NTIRE 2026 team} \\
\textit{Title:} NTIRE 2026 Efficient Low Light Image Enhancement Challenge\\
\textit{Members:} \\
Jiebin Yan\textsuperscript{1} (yanjiebin@jxufe.edu.cn), \\
Chenyu Tu\textsuperscript{1} (2202420256@stu.jxufe.edu.cn), \\
Weixia Zhang\textsuperscript{2} (zwx8981@sjtu.edu.cn), \\
Zhihua Wang\textsuperscript{3} (zhihua.wang@my.cityu.edu.hk), \\
Peibei Cao\textsuperscript{4} (cpb@nuist.edu.cn), \\
Qinghua Lin\textsuperscript{5} (dear.muhua@gmail.com), \\
Yuming Fang\textsuperscript{1} (fa0001ng@e.ntu.edu.sg), \\
Xiaoning Liu\textsuperscript{6} (liuxiaoning2016@sina.com), \\
Zongwei Wu\textsuperscript{7} (zongwei.wu@uni-wuerzburg.de), \\
Zhuyun Zhou\textsuperscript{7} (zhuyun.zhou@uni-wuerzburg.de), \\
Radu Timofte\textsuperscript{7} (radu.timofte@uni-wuerzburg.de) \\
\textit{Affiliations:} \\
\textsuperscript{1} Jiangxi University of Finance and Economics, China \\
\textsuperscript{2} Shanghai Jiao Tong University, China \\
\textsuperscript{3} Sun Yat-sen University, China \\
\textsuperscript{4} Nanjing University of Information Science and Technology, China \\
\textsuperscript{5} Guangdong University of Technology, China \\
\textsuperscript{6} University of Electronic Science and Technology of China, China \\
\textsuperscript{7} Computer Vision Lab, University of Würzburg, Germany
\vspace{1em} % 增加一点垂直间距

\noindent \textbf{MiVideo} \\
\textit{Title:} RetinexFormerRefine: Lightweight RetinexFormer with Spatial and Channel
Attention \\
\textit{Members:} \\
Cheng Li\textsuperscript{1} (licheng8@xiaomi.com), \\
Ziyi Wang\textsuperscript{1}, \\
Peishuai Zha\textsuperscript{1}, \\
Yuqiang Yang\textsuperscript{1}, \\
Jinao Song\textsuperscript{1}, \\
Guangsheng Tang\textsuperscript{1}, \\
Yan Chen\textsuperscript{1}, \\
Long Bao\textsuperscript{1}, \\
Heng Sun\textsuperscript{1} \\
\textit{Affiliations:} \\
\textsuperscript{1} Xiaomi Inc., Beijing, China
\vspace{1em} % 增加一点垂直间距

\noindent \textbf{CVPR\_TCD} \\
\textit{Title:} Grey-guided Cross-Attention Network with FFT Phase Transfer for Efficient Low-Light Image Enhancement\\
\textit{Members:} \\
MD Raqib Khan\textsuperscript{1} (khanmd@tcd.ie), \\
Santosh Kumar Vipparthi\textsuperscript{2} (skvipparthi@iitrpr.ac.in), \\
Subrahmanyam Murala\textsuperscript{1} (muralas@tcd.ie) \\
\textit{Affiliations:} \\
\textsuperscript{1} Trinity College Dublin, The University of Dublin \\
\textsuperscript{2} Indian Institute of Technology Ropar
\vspace{1em} % 增加一点垂直间距

\noindent \textbf{S3} \\
\textit{Title:} Norm U-Net: A Lightweight U-Net with Distribution-Normalizing Preprocessing for Efficient Low-Light Image Enhancement\\
\textit{Members:} \\
Shimon Murai\textsuperscript{1} (shimonmurai@gmail.com), \\
Teppei Kurita\textsuperscript{1} (Teppei.Kurita@sony.com), \\
Ryuta Satoh\textsuperscript{1} (Ryuta.Satoh@sony.com), \\
Yusuke Moriuchi\textsuperscript{1} (Yusuke.Moriuchi@sony.com) \\
\textit{Affiliations:} \\
\textsuperscript{1} Sony Semiconductor Solutions Corporation
\vspace{1em} % 增加一点垂直间距

\noindent \textbf{sun} \\
\textit{Title:} Team sun: A Compact Retinex-Restormer with Six-Metric Checkpoint Selection\\
\textit{Members:} \\
Shang-Quan Sun\textsuperscript{1} (shangquan.sun@ntu.edu.sg), \\
Chenghao Qian\textsuperscript{2} (tscq@leeds.ac.uk), \\
Ruoyu Chen\textsuperscript{3} (cryexplorer@gmail.com), \\
Zhirui Liu\textsuperscript{4} (liuzhr28@mail2.sysu.edu.cn), \\
Wenqi Ren\textsuperscript{4} (renwq3@mail.sysu.edu.cn) \\
\textit{Affiliations:} \\
\textsuperscript{1} Nanyang Technological University, Singapore \\
\textsuperscript{2} University of Leeds, UK \\
\textsuperscript{3} University of Chinese Academy of Sciences, China \\
\textsuperscript{4} Sun Yat-sen University (Shenzhen Campus), China
\vspace{1em} % 增加一点垂直间距

\noindent \textbf{NCHU-CVLab} \\
\textit{Title:} Efficient-HVI: Efficient Low-light Image Enhancement via HVI color space\\
\textit{Members:} \\
Zhidong Zhu\textsuperscript{1} (zhidong96@foxmail.com), \\
Bangshu Xiong\textsuperscript{2} (xiongbs@126.com), \\
Qiaofeng Ou\textsuperscript{2} (ou.qiaofeng@nchu.edu.cn), \\
Zhibo Rao\textsuperscript{2} (raoxi36@foxmail.com), \\
Wei Li\textsuperscript{2} (2250432002@email.szu.edu.cn) \\
\textit{Affiliations:} \\
\textsuperscript{1} Beihang University, China \\
\textsuperscript{2} Nanchang Hangkong University, China
\vspace{1em} % 增加一点垂直间距

\noindent \textbf{NUDT\_DeepIter} \\
\textit{Title:} FinalEfficientEnhancerV2 for Low-Light Enhancement\\
\textit{Members:} \\
Haocheng Wu\textsuperscript{1} (3678535280@qq.com), \\
Changjian Wang\textsuperscript{1} (wangcj@nudt.edu.cn), \\
Xuyao Deng\textsuperscript{1} (dengxuyao@nudt.edu.cn), \\
Kele Xu\textsuperscript{1} (xukelele@nudt.edu.cn), \\
Hui Geng\textsuperscript{1} (gengh666666@163.com), \\
Qisheng Xu\textsuperscript{1} (qishengxu@nudt.edu.cn) \\
\textit{Affiliations:} \\
\textsuperscript{1} National University of Defense Technology, China
\vspace{1em} % 增加一点垂直间距

\noindent \textbf{HIT-LLIE-team} \\
\textit{Title:} MobileIE-6Ch: Efficient Low-Light Image Enhancement for NTIRE 2026\\
\textit{Members:} \\
Xinbai Wang\textsuperscript{1} (2024112824@stu.hit.edu.cn), \\
Duo Liu\textsuperscript{1} (duoliu@stu.hit.edu.cn) \\
\textit{Affiliations:} \\
\textsuperscript{1} Harbin Institute of Technology, Harbin, China
\vspace{1em} % 增加一点垂直间距

\noindent \textbf{VARCHASVI\_SVNIT} \\
\textit{Title:} SCALENet: Scene-Conditioned Adaptive Luminance Enhancement Network\\
\textit{Members:} \\
Arya Shah\textsuperscript{1} (aryashah616@gmail.com), \\
Hriday Kamlesh Samdani\textsuperscript{1} (hridaysamdani2330@gmail.com), \\
Kishor Upla\textsuperscript{1}, \\
Kiran Raja\textsuperscript{2} \\
\textit{Affiliations:} \\
\textsuperscript{1} Sardar Vallabhbhai National Institute of Technology (SVNIT), Surat, India \\
\textsuperscript{2} Norwegian University of Science and Technology (NTNU), Norway
\vspace{1em} % 增加一点垂直间距

\noindent \textbf{Xie\_Liu} \\
\textit{Title:} HE-Net: An Efficient Network for Low-Light Image Enhancement via Histogram Equalization Refinement \\
\textit{Members:} \\
Chen Xie\textsuperscript{1} (2048872092@qq.com), \\
Jie Liu\textsuperscript{1} (blessingbest@163.com) \\
\textit{Affiliations:} \\
\textsuperscript{1} Harbin Institute of Technology, China
\vspace{1em} % 增加一点垂直间距

\noindent \textbf{JialuXu(IVC)} \\
\textit{Title:} JialuXu(IVC)\\
\textit{Members:} \\
Jialu Xu\textsuperscript{1} (j565xu@uwaterloo.ca) \\
\textit{Affiliations:} \\
\textsuperscript{1} IVC Lab, University of Waterloo, Canada
\vspace{1em} % 增加一点垂直间距

\noindent \textbf{Bustaaa} \\
\textit{Title:} Lightweight Dual-Branch Network for Efficient Low-Light Image Enhancement\\
\textit{Members:} \\
Haodian Wang\textsuperscript{1,2} (wanghaodian@mail.ustc.edu.cn) \\
\textit{Affiliations:} \\
\textsuperscript{1} University of Science and Technology of China, China \\
\textsuperscript{2} CHN Energy Digital Intelligence Technology Development (Beijing) Co., Ltd., China
\vspace{1em} % 增加一点垂直间距

\noindent \textbf{sysu701} \\
\textit{Title:} Team sysu701's Technical Report for NTIRE-2026 LLIE\\
\textit{Members:} \\
Hao Zhang\textsuperscript{1} (zhangh758@mail2.sysu.edu.cn), \\
Hongxi Huang\textsuperscript{1} (huanghx93@mail2.sysu.edu.cn), \\
Jinyi Xu\textsuperscript{1} (xujy297@mail2.sysu.edu.cn), \\
Pengfei Wei\textsuperscript{1} (weipf@mail2.sysu.edu.cn), \\
Yanhui Li\textsuperscript{1} (liyh665@mail2.sysu.edu.cn) \\
\textit{Affiliations:} \\
\textsuperscript{1} School of Cyber Science and Technology, Sun Yat-sen University, China
\vspace{1em} % 增加一点垂直间距

\noindent \textbf{SYSU-FVL-ELLIE} \\
\textit{Title:} Lightweight Low-light Ultra-High-Definition Image Enhancement with Hybrid CNN-Attention Network\\
\textit{Members:} \\
Zhi Jin\textsuperscript{1,2,3} (jinzh26@mail.sysu.edu.cn), \\
Le Su\textsuperscript{1} (sule@mail2.sysu.edu.cn), \\
Luwei Tu\textsuperscript{1} (tulw@mail2.sysu.edu.cn), \\
Fei Tan\textsuperscript{1} (tanf29@mail2.sysu.edu.cn), \\
Jiawei Wu\textsuperscript{1} (wujw97@mail2.sysu.edu.cn) \\
\textit{Affiliations:} \\
\textsuperscript{1} School of Intelligent Systems Engineering, Shenzhen Campus of Sun Yat-sen University, Shenzhen, China \\
\textsuperscript{2} Guangdong Provincial Key Laboratory of Fire Science and Intelligent Emergency Technology, Shenzhen, China \\
\textsuperscript{3} Guangdong Provincial Key Laboratory of Robotics and Digital Intelligent Manufacturing Technology, Guangzhou, China
\vspace{1em} % 增加一点垂直间距

\noindent \textbf{KLETech-CEVI} \\
\textit{Title:} ResLNet: Lightweight Residual Network for Low-Light Image Enhancement\\
\textit{Members:} \\
Nikhil Akalwadi\textsuperscript{1,4} (nikhil.akalwadi@kletech.ac.in), \\
Harshitha M Banakar\textsuperscript{1,4} (01fe23bcs193@kletech.ac.in), \\
Sneha S Channappagoudar\textsuperscript{1,4} (01fe23bcs501@kletech.ac.in), \\
Spurti S M K\textsuperscript{1,4} (01fe23bcs286@kletech.ac.in), \\
Tejas Kumar M\textsuperscript{2,4} (01fe23bci052@kletech.ac.in), \\
Ramesh Ashok Tabib\textsuperscript{3,4} (ramesh\_t@kletech.ac.in), \\
Uma Mudenagudi\textsuperscript{3,4} (uma@kletech.ac.in) \\
\textit{Affiliations:} \\
\textsuperscript{1} School of Computer Science and Engineering, KLE Technological University, India \\
\textsuperscript{2} School of Computer Science and Engineering (AI), KLE Technological University, India \\
\textsuperscript{3} Department of Electronics and Communication Engineering, KLE Technological University, India \\
\textsuperscript{4} Center for Visual Intelligence (CEVI), KLE Technological University, Hubballi, Karnataka, India
\vspace{1em} % 增加一点垂直间距

\noindent \textbf{ShinNam!} \\
\textit{Title:} BlindLLIENet: A Lightweight Context-Adaptive Network for Real-World Image Restoration\\
\textit{Members:} \\
Ju-Hyeon Nam\textsuperscript{1} (jhnam0514@inha.edu), \\
Hyemin Shin\textsuperscript{2} (shm9921@keti.re.kr), \\
Sang-Chul Lee\textsuperscript{1} (sclee@inha.ac.kr) \\
\textit{Affiliations:} \\
\textsuperscript{1} Inha University, Republic of Korea \\
\textsuperscript{2} Korea Electronics Technology Institute (KETI), Republic of Korea
\vspace{1em} % 增加一点垂直间距

\noindent \textbf{IIMAS-UNAM} \\
\textit{Title:} IIMAS-UNAM Solution for the NTIRE 2026 Efficient Low-Light Image Enhancement Challenge\\
\textit{Members:} \\
Laura Álvarez-González\textsuperscript{1} (laura.alvargonza@gmail.com), \\
Gibran Fuentes-Pineda\textsuperscript{2} (gibranfp@aries.iimas.unam.mx), \\
Erik Molino-Minero-Re\textsuperscript{2} (erik.molino@iimas.unam.mx) \\
\textit{Affiliations:} \\
\textsuperscript{1} Posgrado en Ciencia e Ingeniería de la Computación, Universidad Nacional Autónoma de México (UNAM), Mexico \\
\textsuperscript{2} Instituto de Investigaciones en Matemáticas Aplicadas y en Sistemas (IIMAS), Universidad Nacional Autónoma de México (UNAM), Mexico
\vspace{1em} % 增加一点垂直间距

\noindent \textbf{Cidaut AI} \\
\textit{Title:} NIEL Solution \\
\textit{Members:} \\
Juan C. Benito\textsuperscript{1} (juaben@cidaut.es), \\
Marcos V. Conde\textsuperscript{2} (marcos.conde@uni-wuerzburg.de), \\
Alvaro Garcia\textsuperscript{1} (alvgar@cidaut.es) \\
\textit{Affiliations:} \\
\textsuperscript{1} Cidaut AI, Spain \\
\textsuperscript{2} Computer Vision Lab, University of Würzburg, Germany
\vspace{1em} % 增加一点垂直间距

\begin{figure}[!htbp]
    \centering
    \includegraphics[width=8cm]{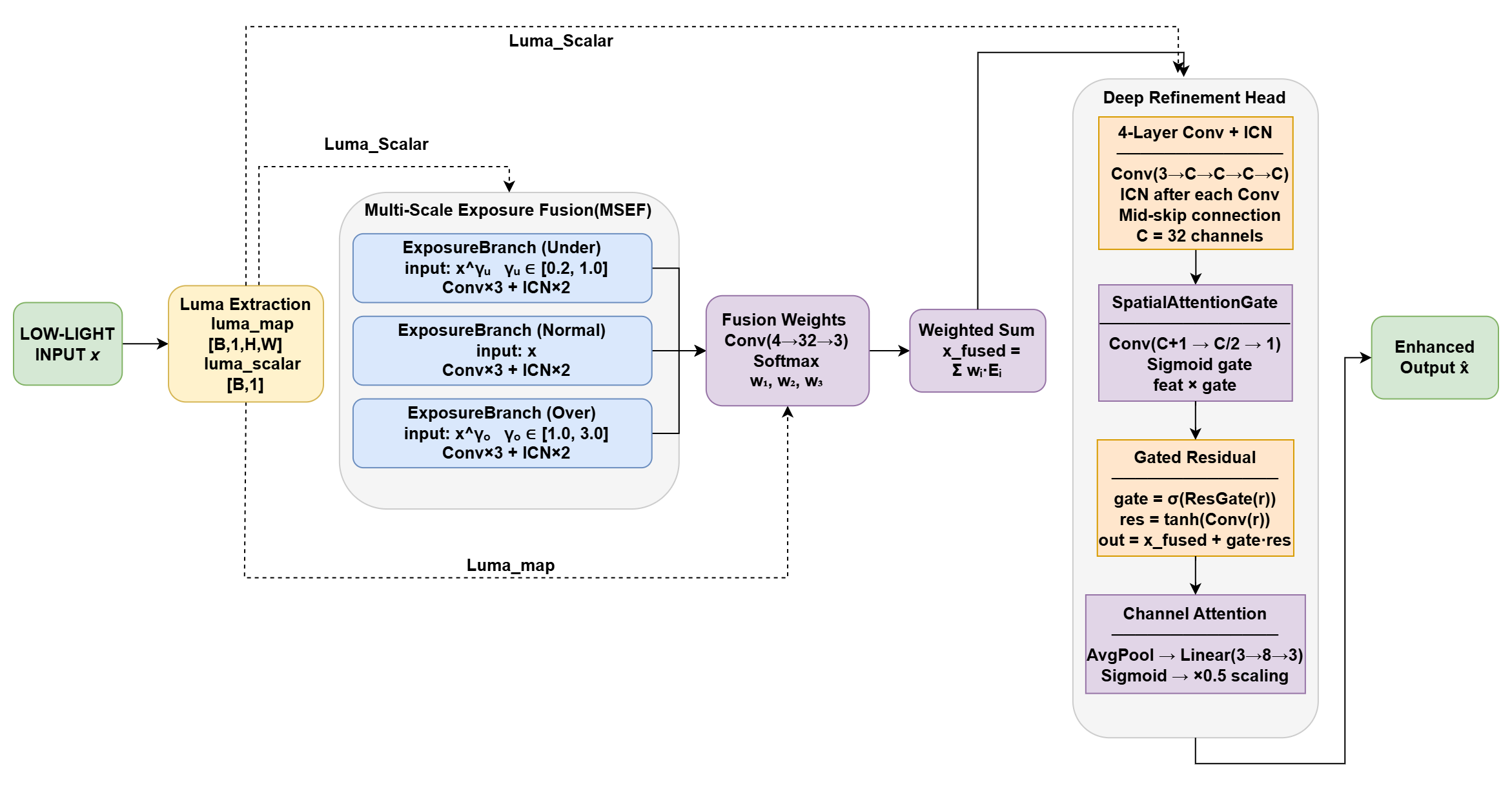}
    % \framebox(230,150){}
    \caption{Architecture of Team VARCHASVI\_SVNIT.}
    \label{fig:VARCHASVI_SVNIT}
\end{figure}

\begin{figure}[!htbp]
    \centering
    \includegraphics[width=8cm]{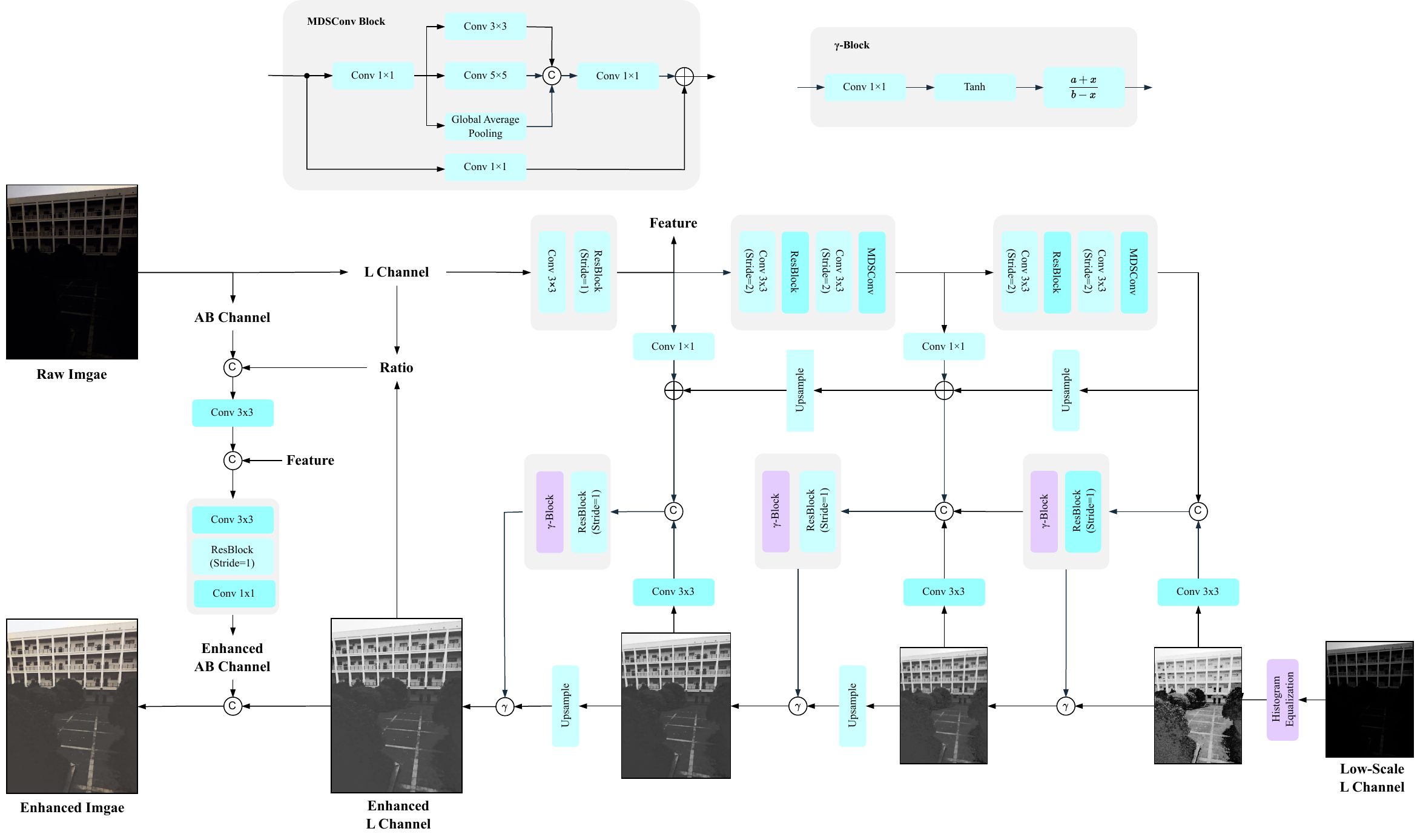}
    % \framebox(230,150){}
    \caption{Architecture of Team Xie\_Liu.}
    \label{fig:xieliu}
\end{figure}

\begin{figure}[!htbp]
    \centering
    \includegraphics[width=8cm]{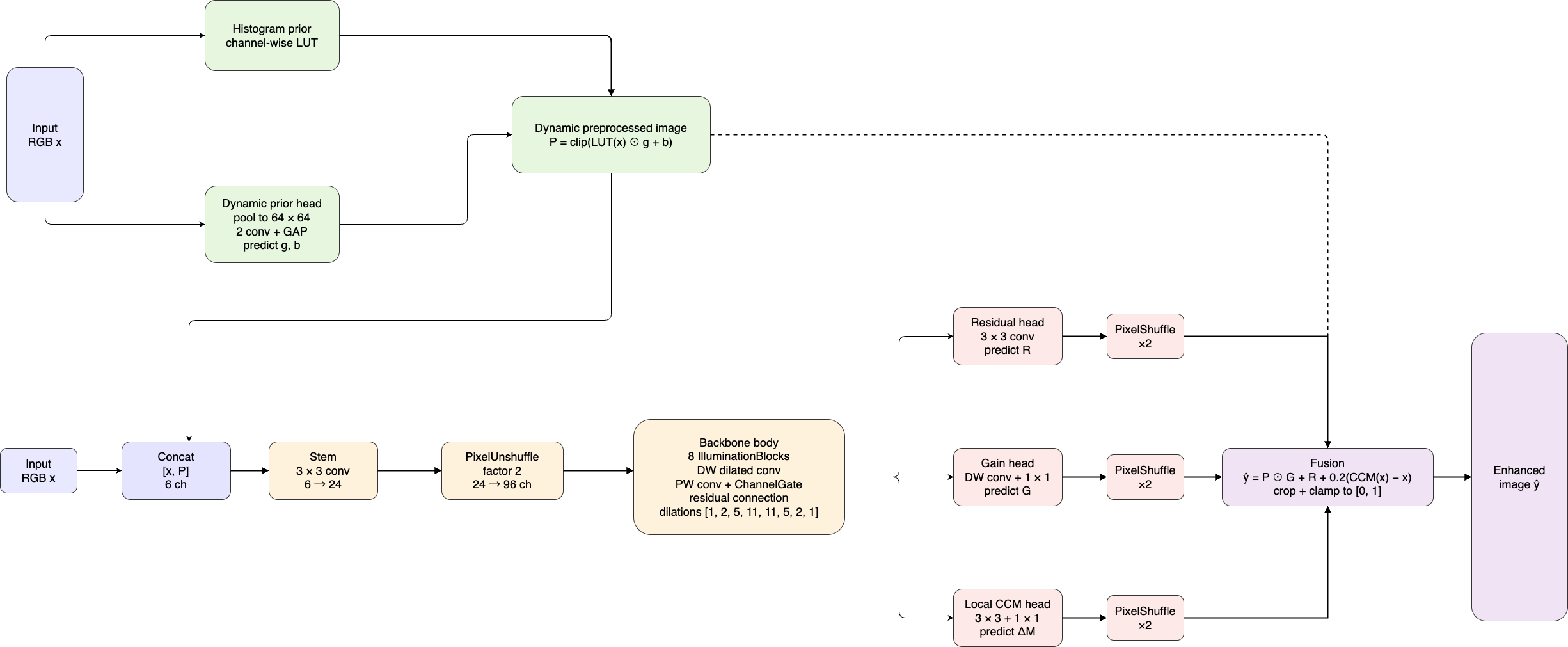}
    % \framebox(230,150){}
    \caption{Architecture of Team JialuXu(IVC).}
    \label{fig:JialuXu}
\end{figure}

\begin{figure}[!htbp]
    \centering
    \includegraphics[width=8cm]{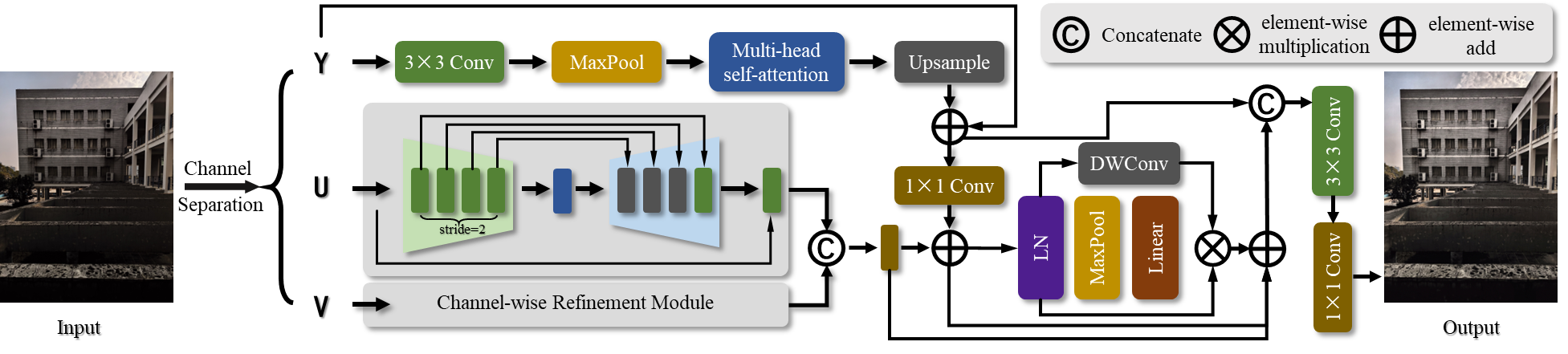}
    % \framebox(230,150){}
    \caption{Architecture of Team Bustaaa.}
    \label{fig:Bustaaa}
\end{figure}

\begin{figure}[!htbp]
    \centering
    \includegraphics[width=8cm]{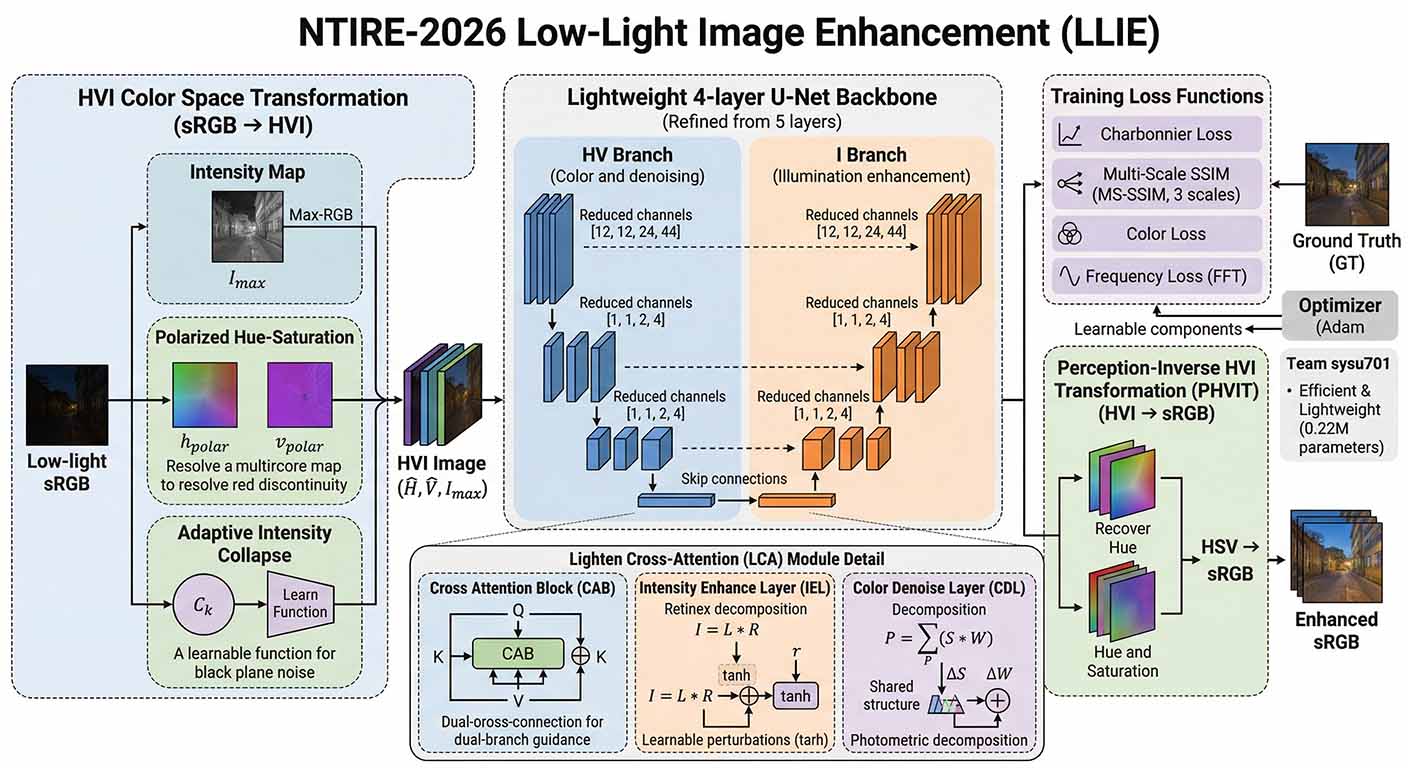}
    % \framebox(230,150){}
    \caption{Architecture of Team sysu\_701.}
    \label{fig:sysu701}
\end{figure}

\begin{figure}[!htbp]
    \centering
    \includegraphics[width=8cm]{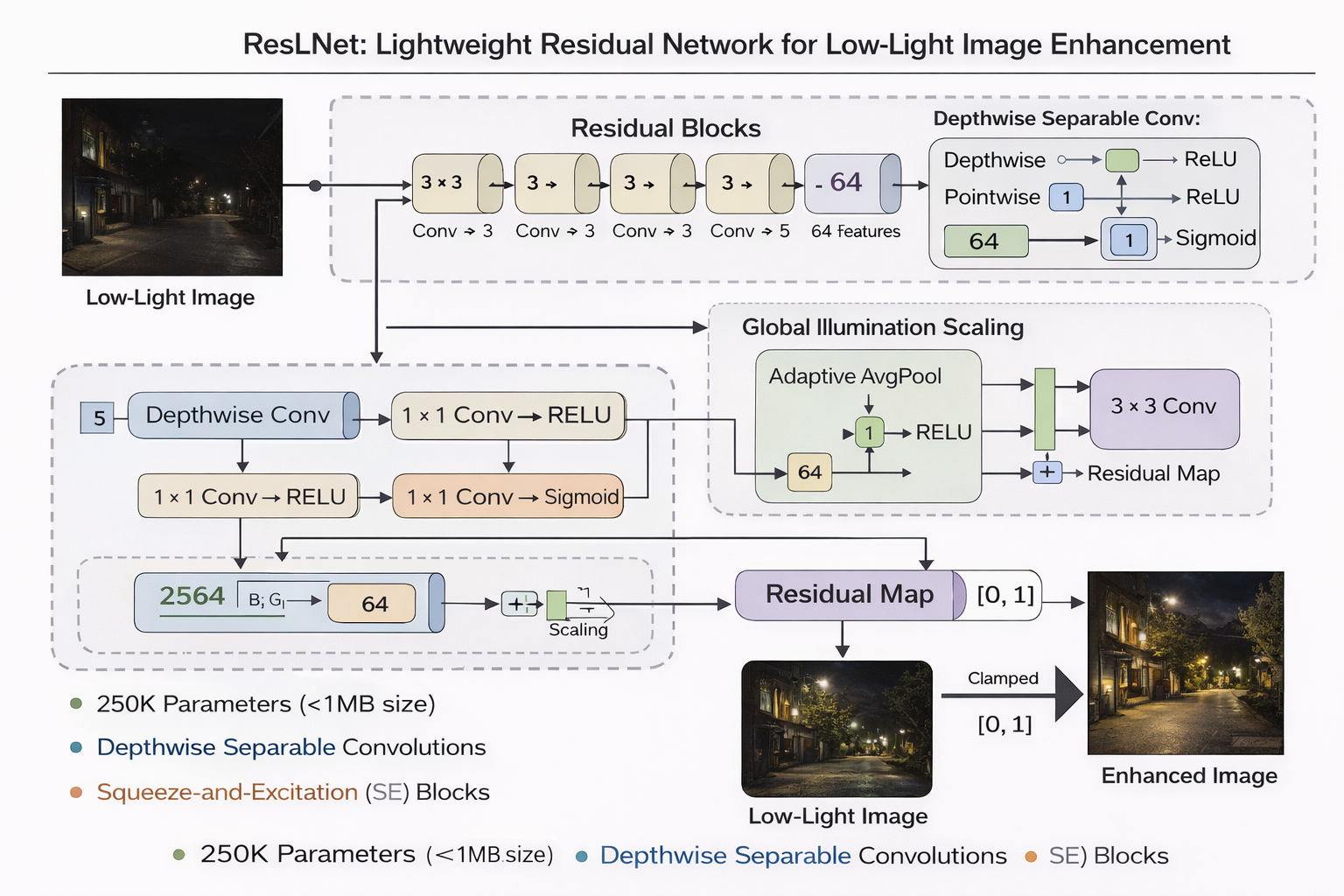}
    % \framebox(230,150){}
    \caption{Architecture of Team KLETech-CEVI.}
    \label{fig:KLETech-CEVI}
\end{figure}

\begin{figure}[!htbp]
    \centering
    \includegraphics[width=8cm]{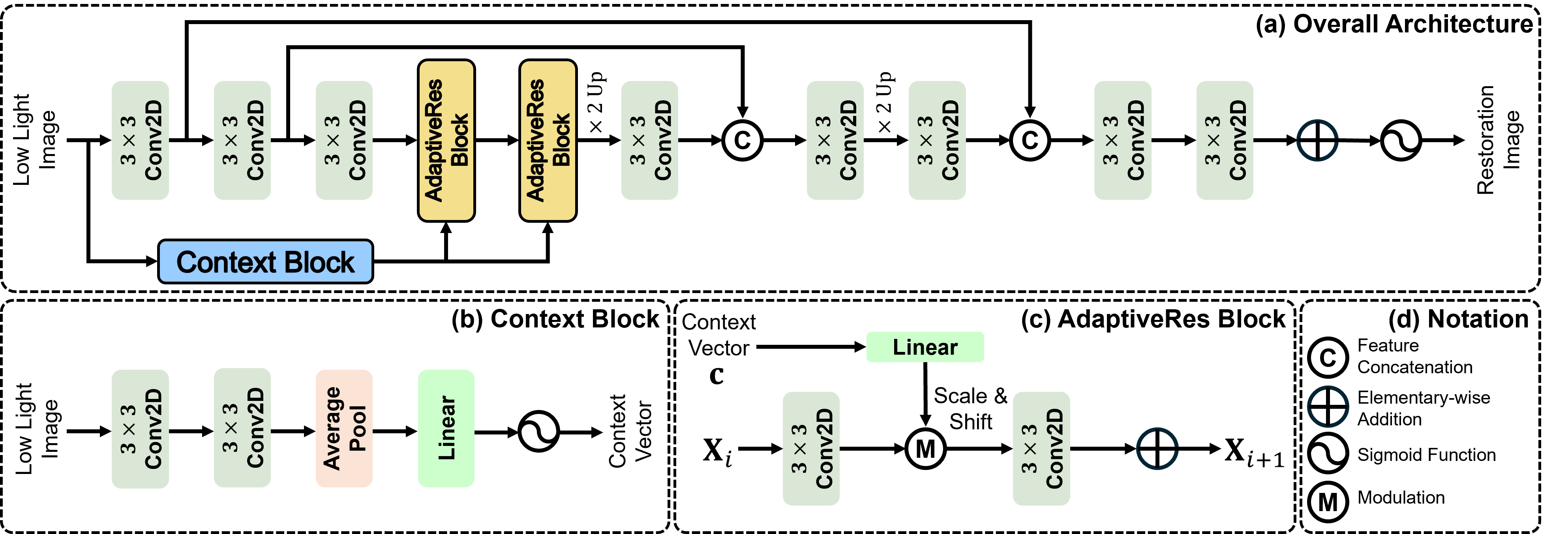}
    % \framebox(230,150){}
    \caption{Architecture of Team ShinNam!.}
    \label{fig:ShinNam!}
\end{figure}

\begin{figure}[!htbp]
    \centering
    \includegraphics[width=8cm]{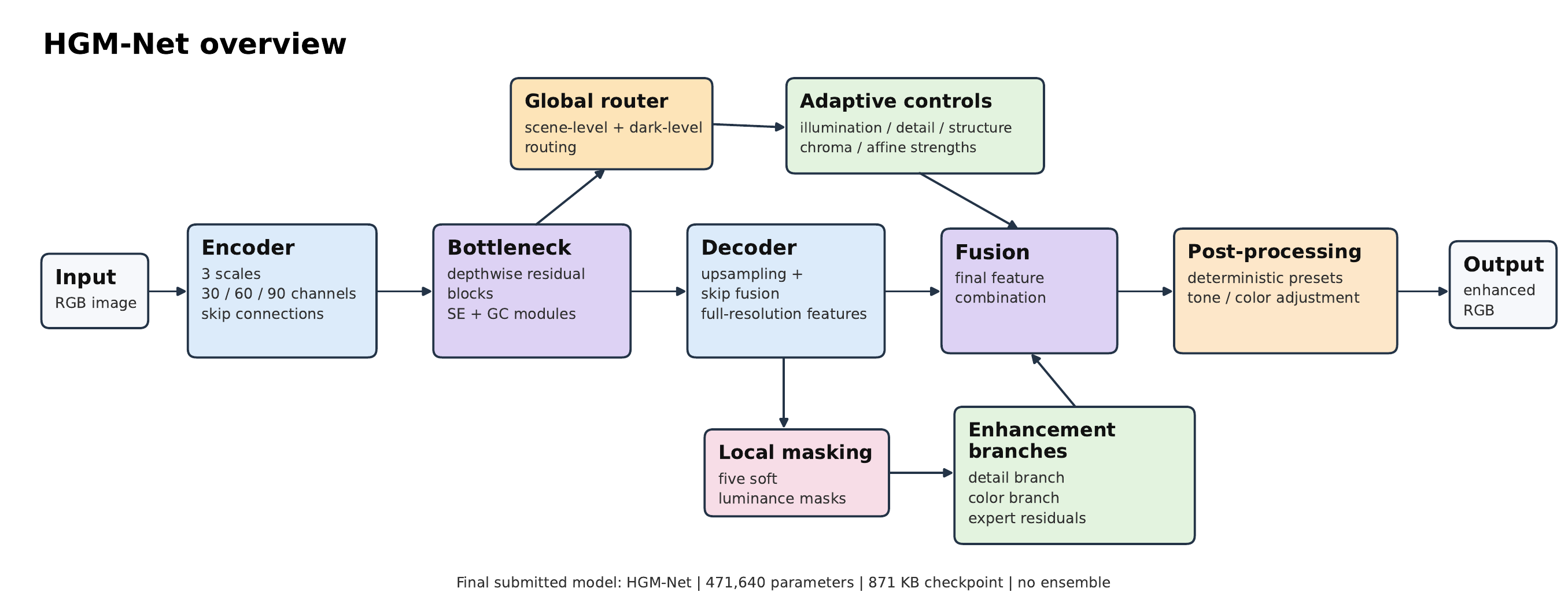}
    % \framebox(230,150){}
    \caption{Architecture of Team IIMAS-UNAM.}
    \label{fig:IIMAS-UNAM}
\end{figure}

\begin{figure}[!htbp]
    \centering
    \includegraphics[width=8cm]{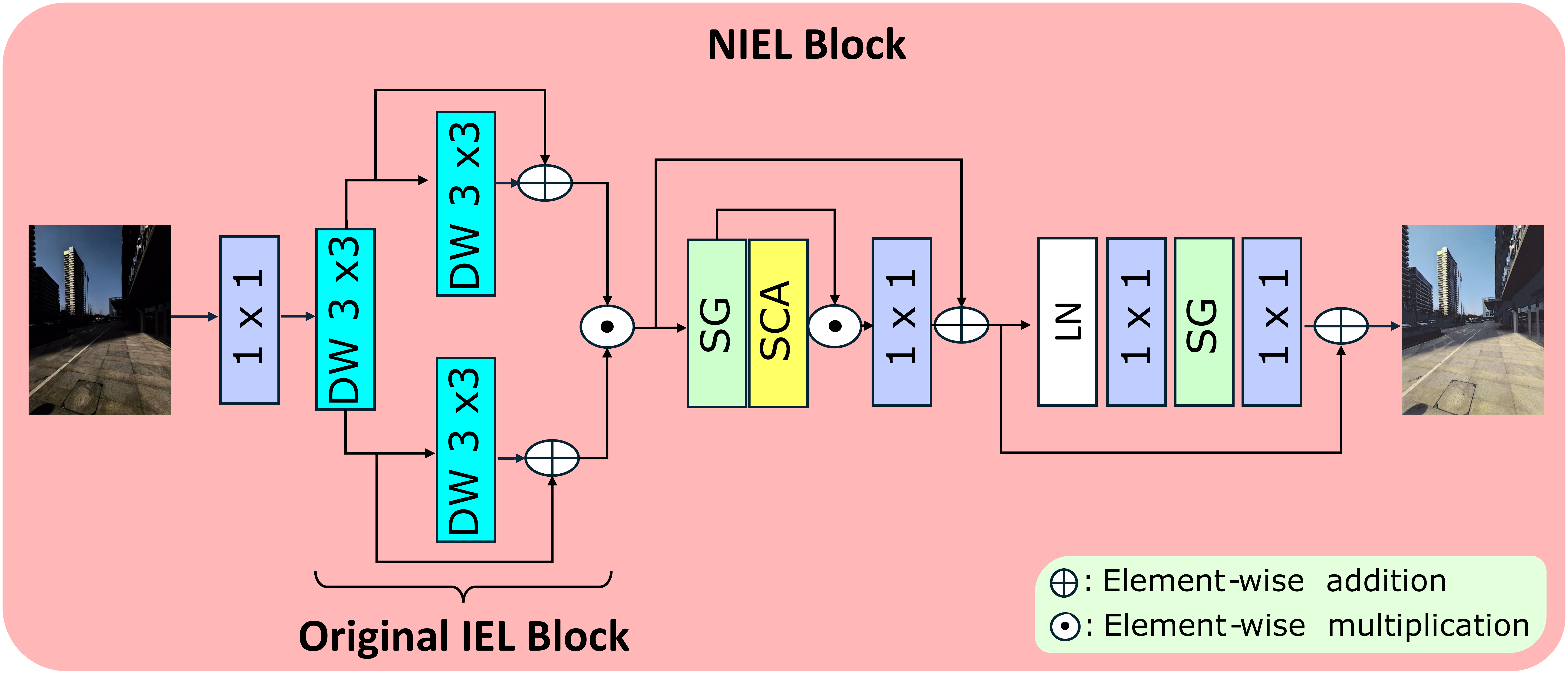}
    % \framebox(230,150){}
    \caption{Architecture of Team Cidaut AI.}
    \label{fig:Cidaut AI}
\end{figure}

\begin{table*}[t]
\centering
\footnotesize
\setlength{\tabcolsep}{3pt}   % 减小列间距（关键）
\renewcommand{\arraystretch}{0.95} % 压缩行高（关键）

\caption{Evaluation and Rankings in the NTIRE 2026 Efficient Low Light Image Enhancement Challenge. Gray indicates groups that participated in the final testing but did not submit a technical report; these groups are therefore omitted from the main text.}
\label{tab:submission_tab}

\resizebox{\textwidth}{!}{%
\begin{tabular}{ccccccccc}
\toprule
\textbf{Team} & \textbf{SSIM (Rank)} & \textbf{LPIPS (Rank)} & \textbf{DISTS (Rank)} & \textbf{LIQE (Rank)} & \textbf{MUSIQ (Rank)} & \textbf{Q-Align (Rank)} & \textbf{Params} & \textbf{Final Rank} \\
\midrule
MiVideo &0.5654 (16)&0.3632 (1)&0.1376 (1)&3.2561 (1)&68.8805 (1)&3.7699 (1)&927,049 &1\\[2pt]
\rowcolor{gray!15}
XJRes &0.5807 (2)&0.4126 (2)&0.2307 (6)&2.1762 (12)&63.5260 (5)&3.3665 (4)&993,014 &2\\[2pt]
CVPR TCD &0.5500 (23)&0.4801 (9)&0.2005 (2)&2.8661 (3)&63.9676 (4)&3.3778 (2)&557,618 &3\\[2pt]
S3 &0.5183 (25)&0.4157 (3)&0.2222 (4)&2.3865 (6)&64.9443 (3)&3.3718 (3)&741,600 &4\\[2pt]
sun &0.5688 (14)&0.4446 (6)&0.2231 (5)&2.2382 (10)&62.4497 (6)&3.2333 (8)&907,414 &5\\[2pt]
\rowcolor{gray!15}
NTR &0.5570 (18)&0.4732 (8)&0.2095 (3)&2.3064 (8)&67.7654 (2)&3.1919 (11)&1,034,247 &6\\[2pt]
NUDT\_DeepIter &0.5211 (24)&0.5031 (12)&0.2317 (7)&2.4856 (4)&61.7693 (10)&3.2415 (7)&897,160 &7\\[2pt]
NCHU-CVLab &0.5557 (19)&0.5045 (13)&0.2542 (13)&2.3933 (5)&62.0570 (9)&3.3015 (5)&957,239 &8\\[2pt]
HIT-LLIE-team &0.5766 (9)&0.5176 (17)&0.2319 (8)&2.2977 (9)&60.3387 (14)&3.2007 (10)&101,922 &9\\[2pt]
Xie\_Liu &0.5551 (20)&0.5171 (15)&0.2791 (16)&2.3798 (7)&62.0840 (8)&3.2130 (9)&913,388 &10\\[2pt]
JialuXu(IVC) &0.5122 (26)&0.5171 (16)&0.2430 (12)&2.6049 (3)&60.6554 (12)&3.2501 (6)&919,594 &11\\[2pt]
\rowcolor{gray!15}
UPT-eLLIE &0.5757 (10)&0.4254 (4)&0.2396 (9)&1.9806 (19)&60.4973 (13)&2.9532 (22)&1,043,743 &12\\[2pt]
VARCHASVI\_SVNIT &0.5575 (17)&0.4408 (5)&0.2414 (11)&1.5017 (26)&62.2430 (7)&3.0040 (16)&890,915 &13\\[2pt]
Bustaaa &0.5509 (21)&0.5143 (14)&0.2397 (10)&2.1875 (11)&61.2102 (11)&2.9807 (20)&205,361 &14\\[2pt]
SYSU-FVL\_ELLIE &0.5819 (1)&0.5561 (21)&0.2804 (17)&2.1494 (15)&55.9899 (19)&3.0264 (15)&994,871 &15\\[2pt]
\rowcolor{gray!15}
UNet\_YYDS &0.5774 (8)&0.5224 (18)&0.2671 (14)&1.8383 (23)&58.6977 (15)&3.0798 (13)&670,544 &16\\[2pt]
sysu\_701 &0.5748 (11)&0.4828 (10)&0.2924 (21)&2.1568 (14)&56.7397 (18)&2.9837 (19)&965,254 &17\\[2pt]
KLETech-CEVI &0.5791 (5)&0.5894 (24)&0.3019 (25)&2.1737 (13)&57.5888 (17)&2.9884 (18)&525,429 &18\\[2pt]
ShinNam! &0.5789 (6)&0.5711 (23)&0.2677 (15)&2.0345 (18)&53.9563 (23)&2.9956 (17)&726,498 &19\\[2pt]
IIMAS-UNAM &0.5504 (22)&0.5536 (20)&0.2805 (18)&2.0970 (16)&57.9082 (16)&3.1247 (12)&890,915 &20\\[2pt]
\rowcolor{gray!15}
CVIR\_Lab &0.5805 (3)&0.5968 (25)&0.2946 (24)&2.0349 (17)&51.5142 (25)&3.0589 (14)&970,566 &21\\[2pt]
\rowcolor{gray!15}
SDUMVP-LLIE &0.5721 (13)&0.4583 (7)&0.2870 (19)&1.6798 (25)&54.5276 (22)&2.9285 (23)&994,871 &22\\[2pt]
\rowcolor{gray!15}
PAI\_Lab &0.5673 (15)&0.4876 (11)&0.2941 (22)&1.8466 (22)&54.5546 (20)&2.8489 (24)&880,583 &23\\[2pt]
\rowcolor{gray!15}
Conqueror &0.5802 (4)&0.5687 (22)&0.2944 (23)&1.8691 (21)&52.1682 (24)&2.9595 (21)&867,224 &24\\[2pt]
Cidaut AI &0.5731 (12)&0.5282 (19)&0.2882 (20)&1.9627 (20)&54.5327 (21)&2.7156 (25)&797,222 &25\\[2pt]
\rowcolor{gray!15}
cvLab &0.5779 (7)&0.6223 (26)&0.3430 (26)&1.8335 (24)&43.9618 (26)&2.6764 (26)&812,663 &26\\[2pt]
\rowcolor{gray!15}
Amrita LLIR &--&--&--&--&--&--&813,420&--\\
\bottomrule
\end{tabular}%
}
\end{table*}

\end{document}